\documentclass{article}

\usepackage[preprint]{neurips_2026}

\usepackage[utf8]{inputenc} 
\usepackage[T1]{fontenc}    
\usepackage{hyperref}       
\usepackage{url}            
\usepackage{booktabs}       
\usepackage{amsfonts}       
\usepackage{nicefrac}       
\usepackage{microtype}      
\usepackage{xcolor}         
\usepackage{tcolorbox}

\usepackage{microtype}
\usepackage{graphicx}
\usepackage{subcaption}
\usepackage{booktabs} 
\usepackage{multirow} 
\usepackage{hyperref}

\usepackage{algorithm}
\usepackage{algpseudocode}

\newcommand{\pmstd}[1]{\,$\pm$\!{\tiny #1}}

\usepackage{amsmath}
\usepackage{amssymb}
\usepackage{mathtools}
\usepackage{amsthm}
\usepackage{bm}
\usepackage[table]{xcolor} 

\usepackage[capitalize,noabbrev]{cleveref}

\theoremstyle{plain}
\newtheorem{theorem}{Theorem}[section]
\newtheorem{proposition}[theorem]{Proposition}

\theoremstyle{definition}

\theoremstyle{remark}

\usepackage[textsize=tiny]{todonotes}

\title{Rethinking Random Transformers as Adaptive Sequence Smoothers for Sleep Staging}

\author{%
  Guisong Liu \\
  School of Biological Science and Medical Engineering \\
  Southeast University \\
  Nanjing, China \\
  \texttt{230258331@seu.edu.cn} \\
  \And
  Xin Gao \\
  University of Bath \\
  Bath, England \\
  \texttt{xin.gao@bath.ac.uk} \\
  \And
  Martin Dresler \\
  Donders Institute for Brain, Cognition and Behaviour \\
  Radboud University Medical Center \\
  Nijmegen, The Netherlands \\
  \texttt{martin.dresler@donders.ru.nl} \\
  \And
  Jiansong Zhang\thanks{Corresponding author} \\
  School of Computer Science \& Software Engineering \\
  Shenzhen University \\
  Shenzhen, China \\
  \texttt{2453103003@mails.szu.edu.cn} \\
  \And
  Pengfei Wei\thanks{Corresponding author} \\
  School of Biological Science and Medical Engineering \\
  Southeast University \\
  Nanjing, China \\
  \texttt{101014012@seu.edu.cn} \\
}

\begin{document}

\maketitle

\begin{abstract}
Automatic sleep staging commonly adopts Transformers under the assumption that they learn complex long-range dependencies. We challenge this view by revealing a neglected property of sleep sequences: strong local temporal continuity. We show that a randomly initialized Transformer, without any training, substantially improves sleep staging performance and consistently outperforms heuristic smoothing. We formalize this effect via a Random Attention Prior Kernel (RAPK), showing that random self-attention acts as an adaptive smoother by balancing global averaging and content-based similarity while preserving stage transitions. Using two metrics, the Local Smoothness Influence Index (LSII) and the Weighted Transition Entropy (WTE), we provide evidence that most performance gains in Transformer-based sleep staging arise from architectural inductive bias rather than parameter learning. Our results suggest that sleep staging can be effectively addressed with structure-driven smoothing mechanisms rather than complex dependency modeling, enabling more efficient and edge-deployable healthcare systems for large-scale physiological monitoring.
\end{abstract}

\section{Introduction}
\label{sec:Introduction}


Sleep is one of the most fundamental physiological activities in humans and plays a critical role in neurological recovery, metabolic regulation, and long-term health \cite{10.5665/sleep.1846}. Consequently, EEG-based automatic sleep staging has become an important and independent research field in physiological signal analysis\cite{phan2022automatic}, serving as the foundation for large-scale sleep monitoring, clinical diagnosis, and sleep modulation systems \cite{10.1093/sleep/zsx139,liu2025sleepmodulationchallengetransitioning,doi:10.1152/physrev.00007.2025}. Unlike conventional recognition tasks, sleep staging is inherently sequential and strongly temporally continuous: neural rhythms and EEG patterns usually evolve gradually across adjacent 30-second epochs rather than changing abruptly \cite{10210638,book}. As a result, determining the sleep stage from a single isolated epoch like other EEG tasks is often insufficient, particularly near ambiguous stage transitions such as N1 onset. This principle is explicitly reflected in the AASM scoring guidelines, where human experts rely heavily on surrounding temporal context to resolve uncertain boundaries \cite{doi:10.5664/jcsm.6576}. Therefore, effective automatic sleep staging systems must integrate both local waveform recognition and temporal context aggregation.

To satisfy this requirement, most modern sleep staging systems adopt a two-stage paradigm: an epoch-level encoder extracts local time--frequency features \cite{Liao_2022,10.3389/fnins.2023.1218072}, followed by a sequence-level model that integrates information across time \cite{7961240,9176741,10210638,10965348,thapa2025multimodal}. 
In recent years, Transformers have become the dominant choice for this sequence modeling stage \cite{9417097,9697331,deng2024lpsgm,10595067,2025sleepdifformer}. 
Their widespread adoption is commonly justified by the assumption that self-attention is uniquely capable of modeling complex, long-range transition dynamics between sleep stages.

\begin{figure*}[!t]
    \centering
    \includegraphics[width=1\textwidth]{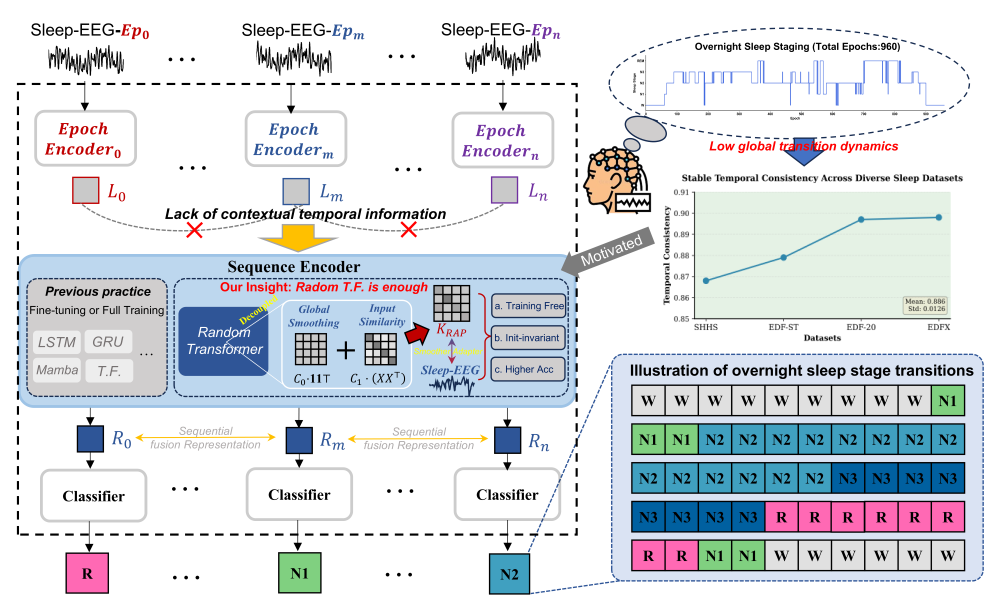}
    \caption{Overview of the proposed framework. Inspired by the low dynamic transition characteristics of sleep staging (right), we propose a Decoupled Random Transformer distinct from conventional fully trained models. This approach utilizes a $K_{RAP}$ adapter to integrate global smoothing and input similarity, enabling efficient, training-free sequence representation fusion and classification.}
    \vspace{-10pt}  
    \label{fig:figure0}
\end{figure*}

Despite this prevailing assumption, empirical observations challenge this narrative.
While Transformer-based architectures consistently achieve state-of-the-art performance in sleep staging \cite{guo_flexsleeptransformer_2024,lee_explainable_2025}, increasing the temporal context length often yields marginal or no performance gains \cite{9697331,10210638,10782964,10782771,10623416}. 
If Transformers truly succeed by capturing long-range dependencies, performance should systematically improve with larger temporal windows. 
This paradox prompts a re-examination of the underlying mechanism:

\begin{tcolorbox}[colback=white, colframe=black, boxrule=0.8pt]
\textbf{What mechanism actually underlies the empirical success of Transformers in sleep staging?}
\end{tcolorbox}

We argue that this paradox can be resolved by revisiting a basic yet underappreciated property of sleep data, its pronounced temporal smoothness. As illustrated in Figure~\ref{fig:figure0}, sleep stages typically persist for extended durations, and genuine transitions occur relatively infrequently. Under such conditions, accurate sequence-level predictions depend less on learning complex transition rules and more on robustly maintaining state continuity while suppressing local noise. From this perspective, the advantage of Transformers may stem not from their ability to learn long-range dependencies, but from an intrinsic architectural bias of self-attention: the aggregation of contextual information that implicitly smooths temporal predictions \cite{10.5555/3295222.3295349}.

Building on this insight, we demonstrate a counter-intuitive phenomenon: a randomly initialized Transformer, without any training, can substantially improve sequence-level sleep staging performance. When stacked on top of a pretrained epoch encoder, this Random Transformer consistently refines temporal predictions, despite containing no learned parameters. Importantly, this effect cannot be explained by naive temporal smoothing. Compared with heuristic smoothers such as moving averages and median filters, which rapidly degrade performance as temporal windows increase, the Random Transformer preserves meaningful stage boundaries while reducing spurious fluctuations. These results indicate that random self-attention functions as an adaptive sequence smoother, one that suppresses noise while selectively retaining structural transitions based on feature similarity.

To explain this behavior, we introduce the Random Attention Prior Kernel (RAPK), a theoretical framework that characterizes the inductive bias of random self-attention. 
Our analysis shows that a random attention layer induces a kernel composed of two complementary components: a global averaging term that promotes temporal smoothness, and a linear similarity term that preserves content-dependent structure. This principled combination enables content-adaptive smoothing to emerge directly from the architecture, even in the absence of learning. 
To empirically examine whether the Transformer acts as a sequence smoother, we introduce two diagnostic metrics: the Local Smoothness Influence Index (LSII) and Weighted Transition Entropy (WTE), measuring local temporal consistency and global transition uncertainty, respectively.Across four public sleep datasets, our results demonstrate that the Transformer enforces temporal continuity(evidenced by higher LSII and lower WTE), effectively acting as a structural regularizer. This smoothing behavior, rather than learned dependency modeling, accounts for the vast majority of the performance gains typically attributed to supervised sequence learning.

Overall, our findings reveal that the success of Transformers in sleep staging is largely driven by architectural priors rather than learned long-range dynamics. This work reframes the role of self-attention in physiological sequence modeling and establishes a strong, training-free baseline that highlights the often overlooked power of inductive bias in structured temporal domains.
The contributions can be summarized as follow:
\begin{enumerate}
    \item We provide the first systematic evidence that a \textbf{Random Transformer}, without any training, significantly improves sequence-level sleep staging across multiple datasets and pretraining paradigms.

    \item We introduce the \textbf{Random Attention Prior Kernel (RAPK)}, providing a theoretical characterization of random self-attention as an \textbf{adaptive sequence smoother} that balances global noise suppression with content-based structure preservation.

    \item We introduce \textbf{LSII} and \textbf{WTE} to quantitatively verify the Transformer's role as a sequence smoother. Our analysis reveals that this intrinsic smoothing bias, rather than learned dependencies, drives the vast majority of performance gains in sleep staging.
\end{enumerate}

\section{Related Work}

\textbf{Sequence Modeling in Sleep Staging.} 
The field has seen a definitive shift from local feature classification to sequence-to-sequence modeling. While early methods utilized Recurrent Neural Networks like LSTMs to capture temporal dependencies~\citep{8631195,lukosevicius_reservoir_2009,seo2020intra}, Transformers have recently established state-of-the-art performance. Contemporary research primarily focuses on enhancing these architectures through complex mechanisms, such as multi-view learning, hierarchical attention, or cross-modal fusion~\citep{10595067,2025sleepdifformer}. These approaches implicitly operate on the premise that the performance ceiling is determined by the model's capacity to learn intricate, long-range dependencies from vast amounts of labeled data. We revisit this fundamental assumption by investigating the intrinsic capabilities of the architecture itself, independent of parameter optimization.

\textbf{Untrained Networks and Structural Priors.} 
The efficacy of random representations is rooted in the seminal concept of \textit{Random Kitchen Sinks}, which demonstrated that fixed, random features can efficiently approximate kernel machines~\citep{10.5555/2981562.2981710}. This principle evolved into \textit{Reservoir Computing} and Echo State Networks, where fixed recurrent weights successfully model complex temporal dynamics~\citep{lukosevicius_reservoir_2009, gallicchio2020deepechostatenetwork}. 
In the deep learning era, the \textit{Deep Image Prior} revealed that the convolutional architecture itself imposes a strong inductive bias for natural image statistics, enabling restoration tasks without any learning~\citep{8579082}. 
While recent works have explored random projections in Transformers to approximate attention mechanisms for efficiency~\citep{choromanski2020rethinking, peng2021randomfeatureattention}, the inductive bias of a fully random self-attention layer remains largely underexplored in physiological time-series. We bridge this gap by establishing the Random Transformer not merely as a feature extractor, but as a specific adaptive smoothing operator tailored for sleep stage continuity.

\textbf{Kernel Interpretations and Low-Pass Filtering.} 
Theoretical frameworks have increasingly linked self-attention to kernel methods, interpreting attention as a non-parametric kernel smoother~\citep{tsai-etal-2019-transformer}. Empirical analyses of \textit{trained} Vision Transformers further suggest that self-attention acts as a low-pass filter, aggregating local information to suppress high-frequency noise~\citep{park2022visiontransformerswork, wang2022antioversmoothingdeepvisiontransformers}. Our work extends these insights to the initialization phase. In contrast to studies that focus on the infinite-width limit for training dynamics, such as the Neural Tangent Kernel or Neural Network Gaussian Process~\citep{10.1145/3406325.3465355, hron2020infiniteattentionnngpntk}, we explicitly characterize the inference dynamics of the random initialization state in the high-dimensional limit via the proposed RAPK framework.We reveal how the interplay of global averaging and linear similarity creates an inherent adaptive smoothing effect that aligns with the structural properties of sleep hypnograms.

\section{Method}

This section analyzes the inductive bias of randomly initialized self-attention through the Random Attention Prior Kernel (RAPK). We show that its high-dimensional structure encodes temporal inertia and boundary preservation, and introduce WTE and LSII to empirically quantify global and local smoothing effects.

\subsection{Random Attention Prior Kernel (RAPK)}
\label{sec:rapk}

We revisit the role of self-attention under random initialization and ask a fundamental question:
what structural bias does self-attention impose on sequences in the absence of learning?
To answer this, we introduce the \emph{Random Attention Prior Kernel} (RAPK),
which captures the intrinsic second-order behavior of randomly initialized self-attention layers
and provides a principled explanation for their sequence-level effects.

\paragraph{Core Insight.}
Random self-attention induces a structured smoothing kernel
that combines global averaging with content-aware similarity.

\begin{proposition}[Structure of RAPK]
\label{prop:rapk}
Let $X \in \mathbb{R}^{T \times d}$ denote a sequence of input representations.
Under random initialization and in the high-dimensional limit, 
the expected Random Attention Prior Kernel satisfies
\begin{equation}
\label{eq:rapk_final}
\mathbb{E}[K_{\mathrm{RAP}}] \approx C_0 \mathbf{1}\mathbf{1}^\top + C_1 XX^\top.
\end{equation}
where $\mathbf{1}$ is the all-ones vector.
The coefficients $C_0$ and $C_1$ are positive, sequence-dependent scalars, whose magnitudes are determined by the initialization variances and sequence length, and scale proportionally with the global statistics of the input features.
\end{proposition}

This decomposition reveals that random self-attention implements a structured smoothing mechanism:
a global low-pass component ($C_0$) that enforces temporal inertia,
and a similarity-preserving component ($C_1$) that maintains discriminative boundaries.

\paragraph{Setup.}
Consider an input sequence of sleep epoch representations $X \in \mathbb{R}^{T \times d}$.
A single attention head is parameterized by projection matrices
$W_Q, W_K, W_V \in \mathbb{R}^{d \times d_k}$,
with independent zero-mean entries and variances
$\sigma_Q^2, \sigma_K^2, \sigma_V^2$, respectively.
The attention output is computed as $O = AV$, where
\begin{equation}
A = \mathrm{Softmax}\!\left( \frac{X W_Q W_K^\top X^\top}{\sqrt{d_k}} \right).
\end{equation}
We define the Random Attention Prior Kernel as $K_{\mathrm{RAP}} \coloneqq O O^\top$.

\paragraph{High-Dimensional and Linear Approximation.}
The derivation relies on two distinct mechanisms. First, standard initialization schemes combined with Layer Normalization naturally constrain the pre-softmax logits ($s_{ip}$) to a narrow range around zero, placing the softmax operation in a near-linear regime and justifying a first-order Taylor expansion. Second, as the projection dimension $d_k$ grows, the Central Limit Theorem and concentration of measure stabilize these high-dimensional inner products, causing the empirical random attention matrix to robustly converge to its expected kernel form (Proposition~3.1). The full derivation is provided in Appendix~\ref{app:rapk_derivation}.

\subsection{Theoretical Alignment with Sleep Physiology}
\label{sec:physio_align}

The significance of Eq.~\ref{eq:rapk_final} extends beyond mathematics; it offers a direct functional mapping to the two fundamental priors of sleep stage transitions. We argue that the success of the Random Transformer stems from the inherent alignment between these mathematical terms and physiological realities.

\paragraph{The Global Term ($C_0$) $\leftrightarrow$ Sleep State Inertia.}
Physiological Prior: Sleep is a continuous, slow-varying process. A fundamental property of sleep hypnograms is \textit{state inertia}: if a subject is in stage N2 at time $t$, they are statistically highly likely to remain in N2 at time $t+1$. Rapid, chaotic fluctuations (e.g., N2 $\to$ W $\to$ N2) within short windows are often artifacts or scoring errors rather than genuine physiological shifts.

Mechanism: The $C_0 \mathbf{1}\mathbf{1}^\top$ term in RAPK enforces this inertia explicitly. By adding a constant bias to the kernel, it promotes a uniform mixing of representations across the temporal window. This acts as a strong structural regularization, dampening high-frequency noise and smoothing out isolated, erroneous predictions produced by the epoch encoder. In essence, $C_0$ embodies the prior belief that "the state is unlikely to change."

\paragraph{The Linear Term ($C_1$) $\leftrightarrow$ Boundary Preservation.}
Physiological Prior: While sleep is stable, transitions between stages (e.g., N2 $\to$ REM) do occur and are biologically meaningful. These transitions are marked by distinct shifts in physiological patterns (e.g., the disappearance of spindles and K-complexes). A valid smoothing algorithm must respect these boundaries; blind averaging would blur them, leading to under-segmentation errors.

Mechanism: The $C_1 XX^\top$ term provides the necessary content-adaptivity. It ensures that the smoothing weight between two epochs $x_i$ and $x_j$ is proportional to their feature similarity $\langle x_i, x_j \rangle$.
\begin{itemize}
    \item \textit{Within a stage:} Features are similar, so $C_1$ reinforces the smoothing effect of $C_0$.
    
    \item \textit{Across a boundary:} Features become dissimilar (orthogonal), minimizing the value of the linear term $C_1 x_i^\top x_j$. This implicitly \textit{gates} the smoothing mechanism: while the global term $C_0$ remains, the loss of the $C_1$ component reduces the aggregate attention weight between distinct stages, thereby preserving the transition boundary.

\end{itemize}
Unlike heuristic smoothers (e.g., Moving Average) that rely solely on temporal proximity, the Random Transformer uses $C_1$ to \textit{sense} the physiological signal, achieving adaptive smoothing that respects state transitions.

\subsection{Metrics for Structural Validation}
\label{sec:metrics}

Standard classification metrics capture prediction accuracy but obscure the temporal mechanisms underlying model behavior.To provide \emph{direct empirical evidence} on the extent to which Transformers, under both trained and random initialization regimes, exhibit sequence-smoothing behavior, we introduce two diagnostic metrics that quantify global transition uncertainty and local temporal consistency. These metrics disentangle architectural smoothing bias from effects induced by learned dependencies.

\paragraph{Weighted Transition Entropy (WTE).}
WTE measures the global stability of the predicted sequence. High-frequency noise manifests as high entropy in the transition matrix. Let $P_{ij}$ be the row-normalized probability of transitioning from stage $i$ to $j$. We define WTE as the class-weighted conditional entropy:
\begin{equation}
    \mathrm{WTE}(S) = \sum_{i=0}^{C-1} \pi_i \left( -\sum_{j} P_{ij} \log P_{ij} \right).
\end{equation}
A lower WTE indicates that the model has successfully suppressed stochastic fluctuations ($C_0$ effect), resulting in a hypnogram with stable, predictable dynamics consistent with healthy sleep architecture.

\paragraph{Local Smoothness Influence Index (LSII).}
LSII measures the local consistency of corrections by assessing whether sequence-level overrides are consistent with neighboring predictions. For the set of corrected indices $\mathcal{C}$, LSII calculates the agreement with the local window $\mathcal{W}_t$:
\begin{equation}
    \mathrm{LSII} = \frac{1}{|\mathcal{C}|} \sum_{t \in \mathcal{C}} \left( \frac{1}{|\mathcal{W}_t|-1} \sum_{k \in \mathcal{W}_t, k \neq t} \mathbb{I}(s_k^{\text{seq}} = s_t^{\text{seq}}) \right).
\end{equation}
A high LSII confirms that the model is actively leveraging local context to resolve ambiguities, validating the adaptive smoothing mechanism. Details of the metrics can be found in \textbf{Appendix~\ref{app:metrics}}.

\section{Experiments}
\label{sec:Experiments}

\subsection{Setup}

\textbf{Datasets.} We evaluated our method on four public sleep EEG datasets: Sleep-EDF-20, Sleep-EDFX, Sleep-EDF-ST, and SHHS~\citep{doi:10.1161/01.CIR.101.23.e215,10.1093/sleep/20.12.1077}. Except for the Generality and Robustness and Metric Analysis sections, all other experiments were conducted on Sleep-EDF-20. Each dataset was split by subject into training, validation, and test sets with an 8:1:1 ratio. Only the first and last 30 minutes of wake epochs were retained to avoid biasing the evaluation metrics. Detailed dataset descriptions are provided in \textbf{Appendix~\ref{app:datasets}}.

\textbf{Epoch Encoder Pretraining.} We adopt iBOT~\citep{zhou2021ibot} as a representative state-of-the-art self-supervised framework to obtain high-quality epoch-level representations. Our conclusions on sequence modeling are \textbf{decoupled} from the specific choice of self-supervision; iBOT is used solely to ensure expressive and robust features. By freezing the encoder, we isolate the sequence modeling stage, enabling a fair assessment of the Random Transformer's intrinsic smoothing behavior independent of feature learning.

\textbf{Implementation Details.} All models were implemented in PyTorch 1.11 and trained on a single NVIDIA RTX 3090 GPU(24GB). We applied no additional signal preprocessing, data augmentation, or class balancing strategies. To ensure reproducibility, each experiment was repeated five times with fixed random seeds ($111\text{--}555$), and results are reported as mean $\pm$ standard deviation. Details of the implementation based on the two-stage sleep staging framework can be found in \textbf{Appendix~\ref{app:Model Details}}.

\subsection{Baseline Comparison}
 After obtaining epoch-level representations from the pretrained encoder, we compared multiple sequence modeling architectures under both random initialization and supervised training to disentangle the performance gains attributable to the architectural inductive bias from those derived from parameter optimization. To rigorously isolate the RAPK's content-dependent term ($C_1$) from naive smoothing, we additionally benchmarked against Moving Average, Median Filter, and a Fixed Attention baseline (uniform $1/W$ weights), which explicitly represents pure global averaging ($C_0$).



\paragraph{Comparison with Sequential Baseline Models.}
To ensure fairness, we evaluate all architectures under two controlled regimes: (i) a random setting without hyperparameter tuning, where all models share identical configurations so that performance differences purely reflect architectural inductive bias, and (ii) a supervised setting with a unified training protocol, including a shared optimizer, early stopping, and minimal architecture-specific tuning. As shown in Table~\ref{tab:baseline}, only the Transformer exhibits substantial performance gains under random initialization, whereas LSTM~\cite{6795963}, GRU~\cite{chung2014empiricalevaluationgatedrecurrent}, TCN~\cite{8099596}, and Mamba~\cite{gu2024mambalineartimesequencemodeling} show no statistically significant improvements. This is expected: recurrent and state-space models rely on parameterized gating or transition dynamics that become unstable under random weights, while TCNs act as random local filters with limited receptive fields, preventing effective sequence-level aggregation.

These observations indicate that the Transformer encoder’s performance gains stem not from random noise or normalization effects, but rather from its inherent inductive bias. Under fully random initialization, the Transformer achieves an ACC of 72.87\%, approximately 4.9\% higher than the non-sequential baseline. Under supervised training, it attains the best sequence modeling performance.Overall, these results suggests that the Transformer encoder can capture sequence dependencies even without supervision.

\begin{table}[t]
\centering
\caption{
Baseline comparison of sequence modeling architectures. \textbf{Random} uses fixed, randomly initialized sequence modules to isolate architectural bias. \textbf{Trained} adds supervised learning. The epoch encoder is pretrained and frozen in all settings. Parentheses show absolute change from the non-sequential baseline (\textbf{None}).}
\label{tab:baseline}

\resizebox{\linewidth}{!}{
\begin{tabular}{lcc|cc}
\toprule
\multirow{2}{*}{Architecture} &
\multicolumn{2}{c|}{Random (Architecture Only)} &
\multicolumn{2}{c}{Trained (Architecture + Optimization)} \\
\cmidrule(lr){2-3} \cmidrule(lr){4-5}
 & ACC (\%) & Weighted F1 (\%) & ACC (\%) & Weighted F1 (\%) \\
\midrule
\rowcolor{gray!20}
None & 67.96 $\pm$ 0.24 & 65.08 $\pm$ 0.29 & -- & -- \\
\midrule
LSTM 
& 41.47 $\pm$ 0.00 {\scriptsize ($\downarrow$26.49)} 
& 24.31 $\pm$ 0.00 {\scriptsize ($\downarrow$40.77)} 
& 74.64 $\pm$ 0.32 {\scriptsize ($\uparrow$6.68)} 
& 73.74 $\pm$ 0.23 {\scriptsize ($\uparrow$8.66)} \\

GRU  
& 61.11 $\pm$ 0.14 {\scriptsize ($\downarrow$6.85)} 
& 56.48 $\pm$ 0.15 {\scriptsize ($\downarrow$8.60)} 
& 74.09 $\pm$ 1.43 {\scriptsize ($\uparrow$6.13)} 
& 73.41 $\pm$ 1.35 {\scriptsize ($\uparrow$8.33)} \\

Mamba 
& 67.50 $\pm$ 0.14 {\scriptsize ($\downarrow$0.46)} 
& 65.22 $\pm$ 0.19 {\scriptsize ($\uparrow$0.14)} 
& 73.34 $\pm$ 1.00 {\scriptsize ($\uparrow$5.38)} 
& 72.37 $\pm$ 1.09 {\scriptsize ($\uparrow$7.29)} \\

TCN 
& 68.05 $\pm$ 0.17 {\scriptsize ($\uparrow$0.09)} 
& 65.25 $\pm$ 0.22 {\scriptsize ($\uparrow$0.17)} 
& 75.36 $\pm$ 0.32 {\scriptsize ($\uparrow$7.40)} 
& 74.08 $\pm$ 0.36 {\scriptsize ($\uparrow$9.00)} \\

\rowcolor{blue!10}
\textbf{Transformer Encoder} 
& \textbf{72.87 $\pm$ 0.25 {\scriptsize ($\uparrow$4.91)}} 
& \textbf{70.55 $\pm$ 0.38 {\scriptsize ($\uparrow$5.47)}} 
& \textbf{77.29 $\pm$ 0.35 {\scriptsize ($\uparrow$9.33)}} 
& \textbf{75.90 $\pm$ 0.48 {\scriptsize ($\uparrow$10.82)}} \\
\bottomrule
\end{tabular}
}
\end{table}

\paragraph{Comparison with Blind Smoothing Baselines.}

\begin{figure*}[t]
    \centering
    \begin{subfigure}{0.80\linewidth}
        \centering
        \includegraphics[width=\linewidth]{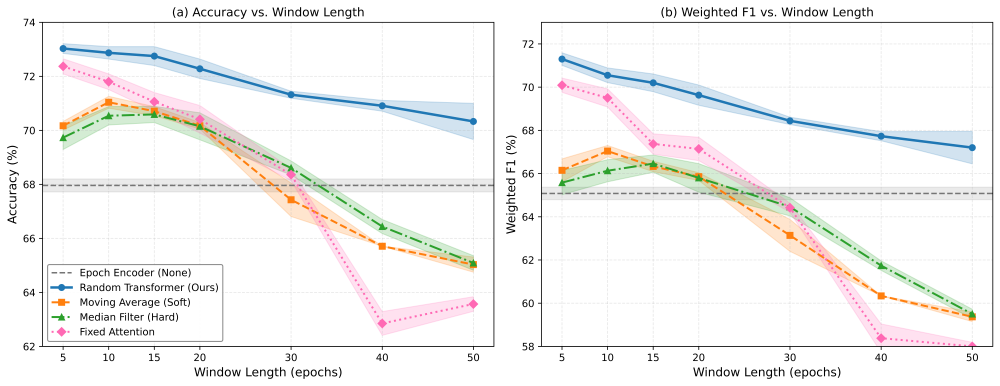}
        \label{fig:window_length}
    \end{subfigure}

    \caption{Sensitivity to window length} 
    \vspace{-10pt} 
    \label{fig:temporal_window_sensitivity}
\end{figure*}

As shown in Figure~\ref{fig:temporal_window_sensitivity}, while heuristic smoothers (Moving Average, Median Filter) and Fixed Attention exploit local continuity at short intervals, their performance collapses catastrophically as the window expands, falling significantly below the non-sequential baseline. This confirms that indiscriminate averaging dilutes discriminative signals (over-smoothing).
In sharp contrast, the Random Transformer maintains robust performance even at $W=50$. This substantial performance gap serves as empirical proof that the linear similarity term ($C_1 XX^\top$) is active: unlike blind filters, the random kernel leverages input geometry to perform content-adaptive smoothing, selectively aggregating context to suppress noise while maintaining distinct stage transitions.

\subsection{Generality and Robustness}
We conducted comprehensive evaluations on four public datasets under two training paradigms: self-supervised pretraining and fully supervised learning. A key observation arises in the fully supervised setting: As shown in Table~\ref{tab:generality}, \emph{when the epoch encoder is trained in a supervised manner and produces highly discriminative representations, further optimization of the subsequent Transformer-based sequence encoder yields negligible, and in some cases no, performance improvements.} Notably, on Sleep-EDF-20, the frozen Random Transformer achieves an accuracy of 79.96\%, surpassing the fully trained sequence model, which attains 78.74\%.

These results suggest that once high-quality local features are established, the remaining sequence modeling problem in sleep staging is dominated by local temporal smoothing. This role can be effectively fulfilled by the inherent inductive bias of random self-attention, without the need for parameter learning. While sequence-level training remains beneficial when epoch representations are imperfect, such as in self-supervised settings, our findings indicate that with sufficiently strong epoch-level encoders, the Random Transformer functions as an adequate, training-free sequence smoother, making additional parameter optimization largely redundant.

\begin{table}[t]
\centering
\scriptsize
\setlength{\tabcolsep}{3.5pt}
\renewcommand{\arraystretch}{0.95}

\caption{
Generality and robustness under different training regimes (ACC / Weighted F1 \%). 
\textbf{Top:} random vs supervised sequence encoders with frozen epoch encoder, assessing sequence-level generalization. 
\textbf{Middle:} fixed random sequence encoder with trained epoch encoder, testing effectiveness of random sequence modeling under supervised learning. 
\textbf{Bottom:} fully supervised end-to-end training, representing upper-bound performance.
}
\label{tab:generality}

\resizebox{\linewidth}{!}{
\begin{tabular}{lcccc}
\toprule
\textbf{Method} & EDF-20 & EDF-ST & EDFX & SHHS \\
\multicolumn{1}{c}{} & \multicolumn{1}{c}{\tiny ACC / F1} &
\multicolumn{1}{c}{\tiny ACC / F1} &
\multicolumn{1}{c}{\tiny ACC / F1} &
\multicolumn{1}{c}{\tiny ACC / F1} \\
\midrule

\multicolumn{5}{l}{\textbf{Random vs. Trained Transformer}} \\

\rowcolor{gray!20}
None &
67.96\pmstd{0.24} / 65.08\pmstd{0.29} &
65.76\pmstd{0.23} / 63.29\pmstd{0.28} &
73.94\pmstd{0.14} / 71.89\pmstd{0.13} &
71.42\pmstd{0.30} / 70.38\pmstd{0.30} \\

\rowcolor{blue!15}
\textbf{Random Transformer} &
\textbf{72.87\pmstd{0.25} / 70.55\pmstd{0.38}} &
\textbf{69.62\pmstd{0.23} / 67.34\pmstd{0.27}} &
\textbf{77.75\pmstd{0.11} / 76.32\pmstd{0.21}} &
\textbf{74.18\pmstd{0.20} / 73.32\pmstd{0.20}} \\

Supervised Transformer &
77.29\pmstd{0.35} / 75.90\pmstd{0.48} &
68.05\pmstd{1.99} / 66.95\pmstd{1.68} &
79.45\pmstd{1.45} / 78.06\pmstd{1.42} &
80.86\pmstd{0.40} / 80.62\pmstd{0.33} \\

\midrule
\multicolumn{5}{l}{\textbf{Sequence-level Frozen Encoder}} \\

One-stage supervised &
75.69\pmstd{0.74} / 74.60\pmstd{0.93} &
74.22\pmstd{0.82} / 73.83\pmstd{0.97} &
76.77\pmstd{0.37} / 75.33\pmstd{0.23} &
80.96\pmstd{0.28} / 79.81\pmstd{0.28} \\

\rowcolor{blue!15}
\textbf{Frozen (w/o Positional Encoding)} &
\textbf{79.96\pmstd{0.42} / 79.13\pmstd{0.73}} &
\textbf{75.64\pmstd{0.79} / 75.28\pmstd{0.84}} &
\textbf{78.67\pmstd{0.79} / 77.83\pmstd{0.64}} &
\textbf{83.36\pmstd{0.14} / 82.89\pmstd{0.18}} \\

\rowcolor{blue!15}
\textbf{Frozen (w/ Positional Encoding)} &
\textbf{79.41\pmstd{0.31} / 78.82\pmstd{0.40}} &
\textbf{74.00\pmstd{2.03} / 73.75\pmstd{2.08}} &
\textbf{78.82\pmstd{0.31} / 78.03\pmstd{0.59}} &
\textbf{83.21\pmstd{0.37} / 82.78\pmstd{0.44}} \\

\midrule
\multicolumn{5}{l}{\textbf{Sequence-level Supervised Encoder}} \\

Two-stage supervised (w/o Positional Encoding) &
79.45\pmstd{0.74} / 78.57\pmstd{0.80} &
75.15\pmstd{1.04} / 74.51\pmstd{1.15} &
79.15\pmstd{0.40} / 78.21\pmstd{0.39} &
83.79\pmstd{0.16} / 83.41\pmstd{0.21} \\

Two-stage supervised (w/ Positional Encoding) &
78.74\pmstd{1.02} / 77.71\pmstd{1.32} &
75.56\pmstd{0.54} / 74.97\pmstd{0.77} &
79.67\pmstd{0.64} / 78.16\pmstd{0.64} &
83.82\pmstd{0.31} / 83.41\pmstd{0.41} \\

\bottomrule
\end{tabular}
}
\end{table}

\subsection{Convergence of the Empirical Random Attention Kernel}
\label{sec:dk_sensitivity}

\begin{figure*}[t]
    \centering
    \begin{subfigure}{0.40\linewidth}
        \centering
        \includegraphics[width=\linewidth]{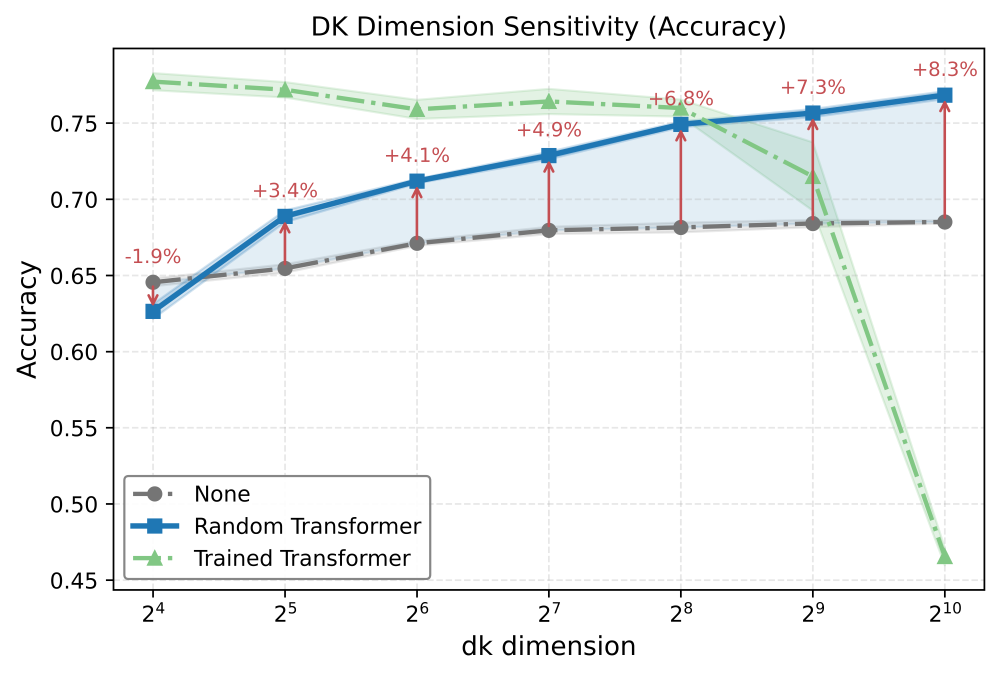}
        \caption{Accuracy Sensitivity}
        \label{fig:dk_acc}
    \end{subfigure}
    \hspace{0.01\linewidth}
    \begin{subfigure}{0.40\linewidth}
        \centering
        \includegraphics[width=\linewidth]{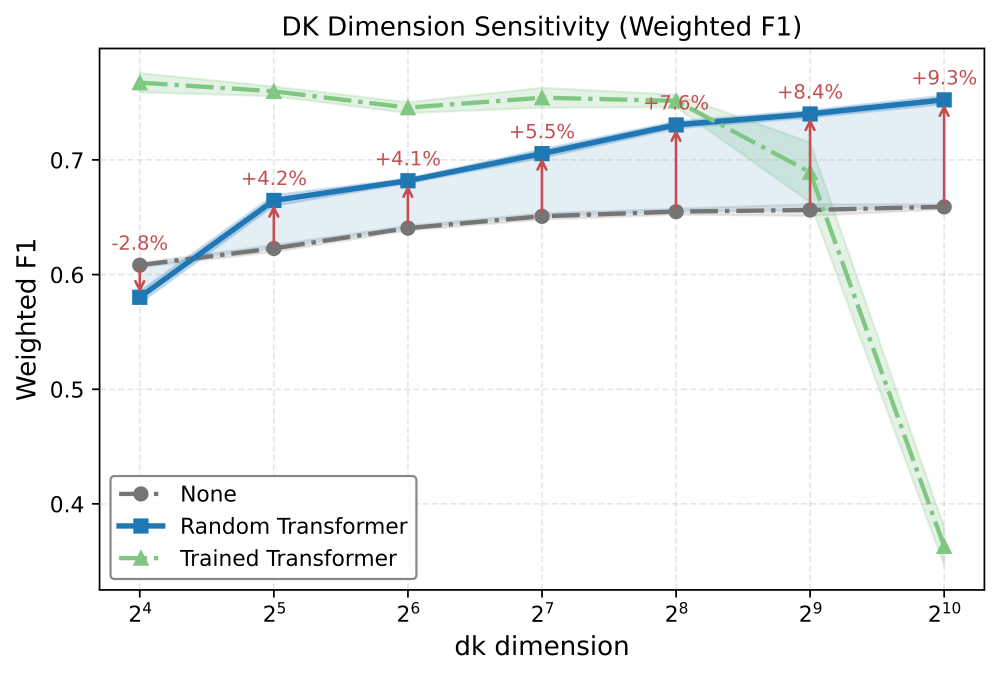}
        \caption{Weighted F1 Sensitivity}
        \label{fig:dk_f1}
    \end{subfigure}
    \caption{Sensitivity analysis of projection dimension ($d_k$).}
    \vspace{-5pt} 
    \label{fig:dk_sensitivity}
\end{figure*}

To explicitly validate the high-dimensional asymptotic assumptions underlying the RAPK theory, we analyze the effect of the projection dimension $d_k$ in the Random Transformer. We vary $d_k$ from 16 to 1024 while keeping the pretrained epoch encoder fixed. The results, shown in Figure~\ref{fig:dk_sensitivity}, reveal fundamentally different scaling behaviors between random and trained Transformers.

As $d_k$ increases, the Random Transformer exhibits a clear and monotonic performance improvement. At low dimensions ($d_k=16$), performance is slightly suppressed due to the information bottleneck and the high variance of low-dimensional random projections. However, \emph{as $\boldsymbol{d_k}$ scales to 1024, accuracy steadily improves, reaching a peak gain of +8.3\% over the non-sequential baseline}. This behavior directly corroborates the RAPK analysis: increasing $d_k$ yields a lower-variance empirical approximation of the expected attention kernel $E[K_{\text{RAP}}]$, allowing the linear similarity term ($C_1 XX^\top$) to dominate random fluctuations while remaining balanced by the global averaging component ($C_0$). In contrast, while the trained Transformer performs competitively at low-to-medium dimensions, it degrades in the high-dimensional regime due to optimization instability and overfitting, illustrating the increasing difficulty of learning in an exploding parameter space.

These results highlight a key advantage of the Random Transformer: \emph{it decouples representational capacity from optimization complexity}. By leveraging high-dimensional random attention, the model can enforce strong, adaptive temporal smoothing without suffering from optimization-induced instability. Together with the baseline comparisons in Table~\ref{tab:baseline}, this experiment provides strong empirical evidence for the RAPK theory, demonstrating that the observed sequence modeling gains arise from the intrinsic high-dimensional kernel structure of random self-attention.

\subsection{Metric Analysis}
We empirically validate the effectiveness of WTE and LSII as diagnostics for sequence modeling behavior. As shown in Table~\ref{tab:metrics}, the non-sequential baseline exhibits high WTE, reflecting fragmented and noisy predictions. Introducing random self-attention consistently reduces WTE and increases LSII across all datasets, demonstrating that the Transformer architecture intrinsically enforces temporal smoothness even without training.

Crucially, the superior performance achieved under supervised training provides direct evidence that \emph{optimizing Transformer-based sequence models effectively amounts to maximizing temporal smoothness.} In this view, training does not introduce a qualitatively different mechanism, but rather aligns model parameters with the same smoothing prior already present in the architecture. In certain datasets (e.g., EDF-ST), supervised optimization slightly degrades WTE and LSII, indicating that the inductive bias of the random structure can, in some cases, enforce smoother sequence behavior than learned projections.Overall, these results underscore the central role of structural priors in shaping the temporal regularization properties of Transformer-based sequence models.








\begin{table}[t]
\centering
\caption{WTE and LSII metrics across datasets. Lower WTE is better ($\downarrow$); higher LSII is better ($\uparrow$). Values in parentheses report absolute changes relative to the specified baseline.}
\label{tab:metrics}

\small
\setlength{\tabcolsep}{5pt}
\renewcommand{\arraystretch}{1.1}

\begin{tabular}{l l c c c c}
\toprule
Metric & Category & EDF-20 & EDF-ST & EDFX & SHHS \\
\midrule

\rowcolor{blue!10}
WTE $\downarrow$ & None 
& 0.873 & 0.806 & 0.737 & 0.793 \\

\rowcolor{blue!20}
WTE $\downarrow$ & Random 
& 0.575 {\tiny($\downarrow$0.30)}
& 0.465 {\tiny($\downarrow$0.34)}
& 0.477 {\tiny($\downarrow$0.26)}
& 0.580 {\tiny($\downarrow$0.21)} \\

\rowcolor{blue!30}
WTE $\downarrow$ & Trained 
& 0.473 {\tiny($\downarrow$0.40)}
& 0.573 {\tiny($\downarrow$0.23)}
& 0.337 {\tiny($\downarrow$0.40)}
& 0.450 {\tiny($\downarrow$0.34)} \\

\rowcolor{blue!40}
WTE $\downarrow$ & True Label 
& 0.404 {\tiny($\downarrow$0.47)}
& 0.407 {\tiny($\downarrow$0.40)}
& 0.329 {\tiny($\downarrow$0.41)}
& 0.451 {\tiny($\downarrow$0.34)} \\

\midrule

\rowcolor{orange!10}
LSII $\uparrow$ & None $\rightarrow$ Random 
& 0.745 & 0.841 & 0.698 & 0.763 \\

\rowcolor{orange!20}
LSII $\uparrow$ & None $\rightarrow$ Trained 
& 0.837 {\tiny($\uparrow$0.09)}
& 0.760 {\tiny($\downarrow$0.08)}
& 0.849 {\tiny($\uparrow$0.15)}
& 0.809 {\tiny($\uparrow$0.05)} \\

\bottomrule
\end{tabular}
\end{table}

\subsection{Qualitative Analysis of Sleep Stage Transitions}
\begin{figure}[t]
\centering

\begin{subfigure}[b]{0.24\columnwidth}
\centering
\includegraphics[width=\textwidth]{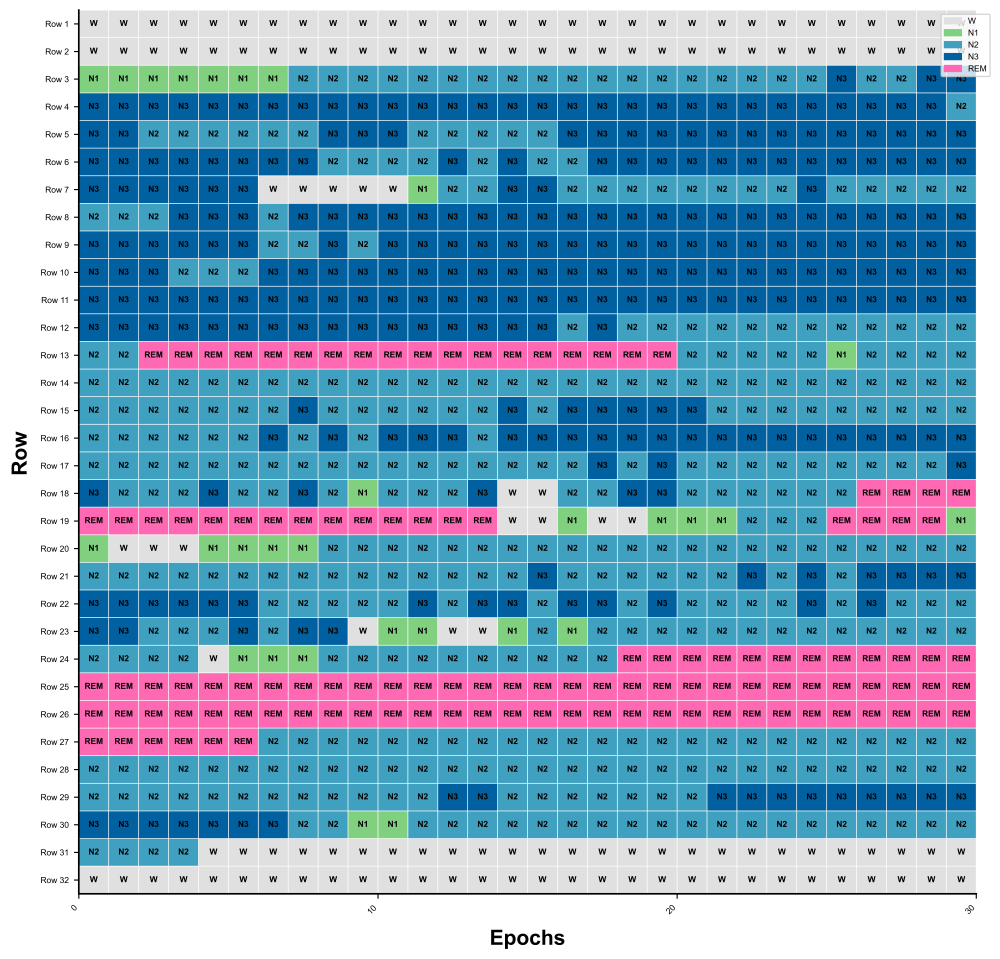}
\caption{Ground Truth}
\label{fig:vis_c}
\end{subfigure}
\hfill
\begin{subfigure}[b]{0.24\columnwidth}
\centering
\includegraphics[width=\textwidth]{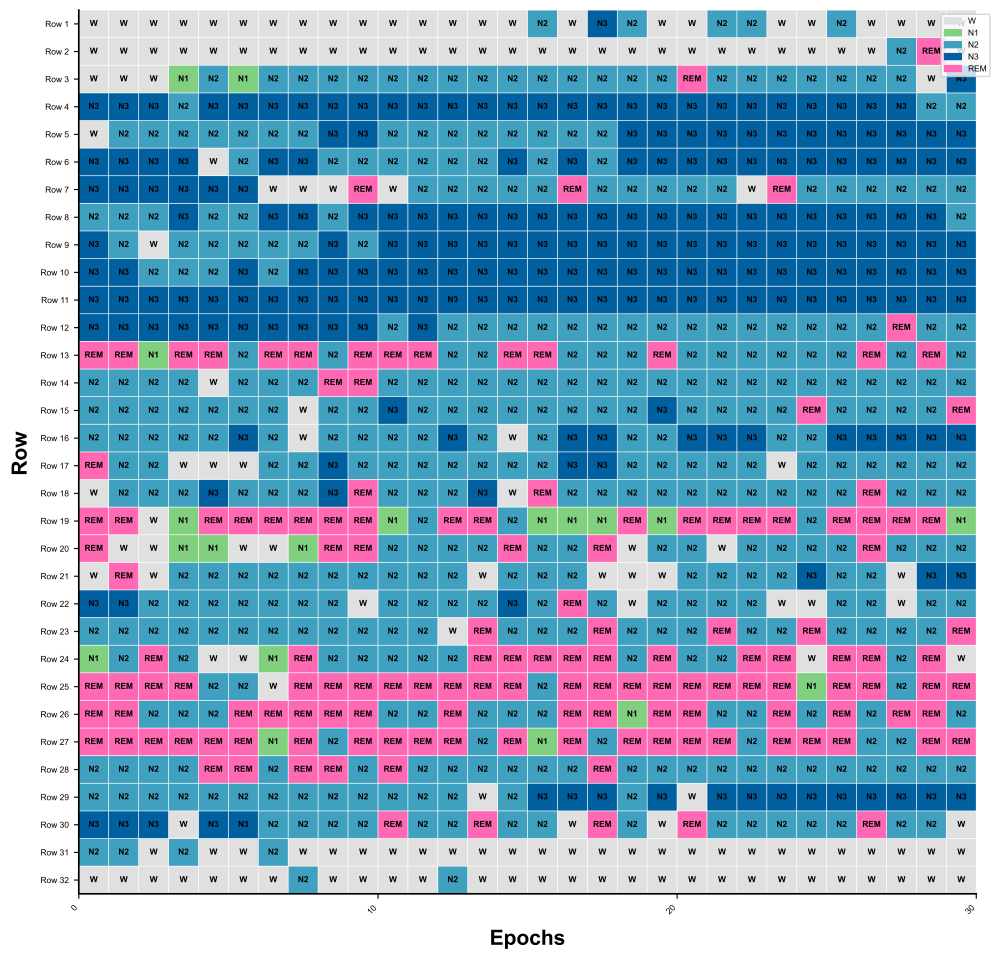}
\caption{Epoch Encoder}
\label{fig:vis_a}
\end{subfigure}
\hfill
\begin{subfigure}[b]{0.24\columnwidth}
\centering
\includegraphics[width=\textwidth]{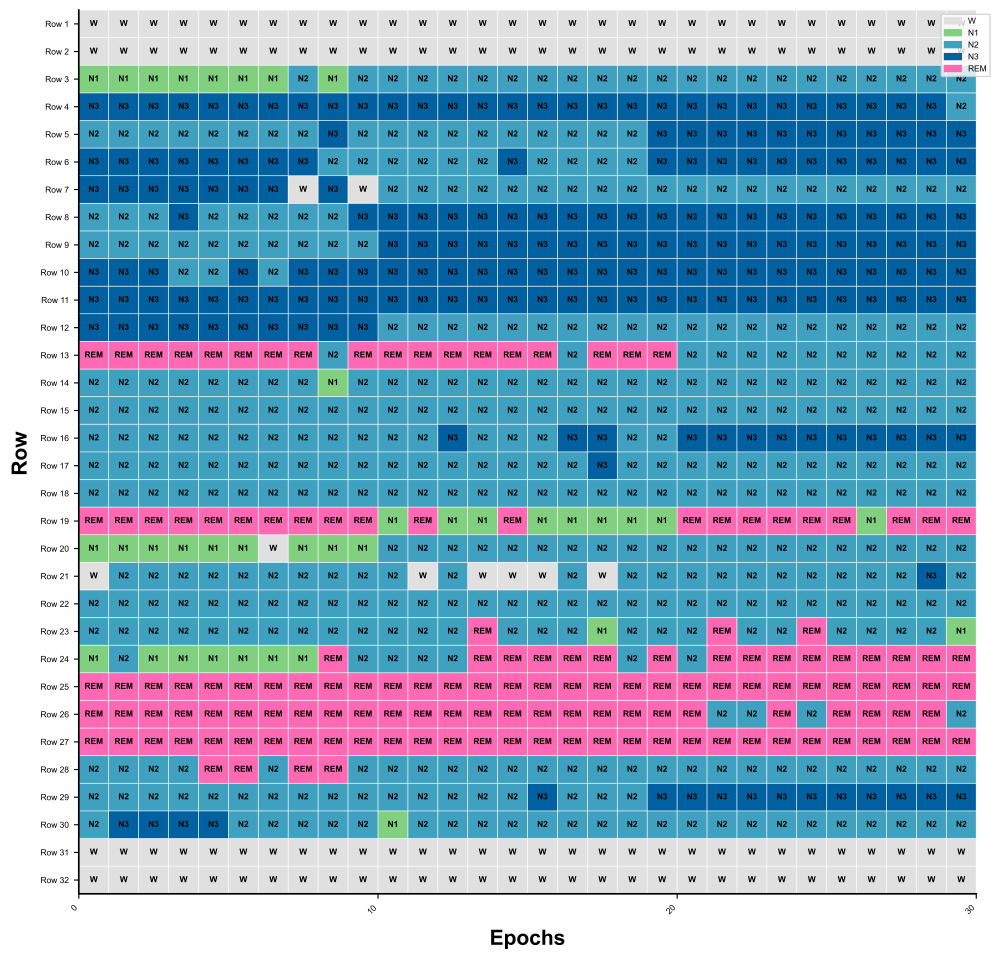}
\caption{Random Transformer}
\label{fig:vis_b}
\end{subfigure}
\hfill
\begin{subfigure}[b]{0.24\columnwidth}
\centering
\includegraphics[width=\textwidth]{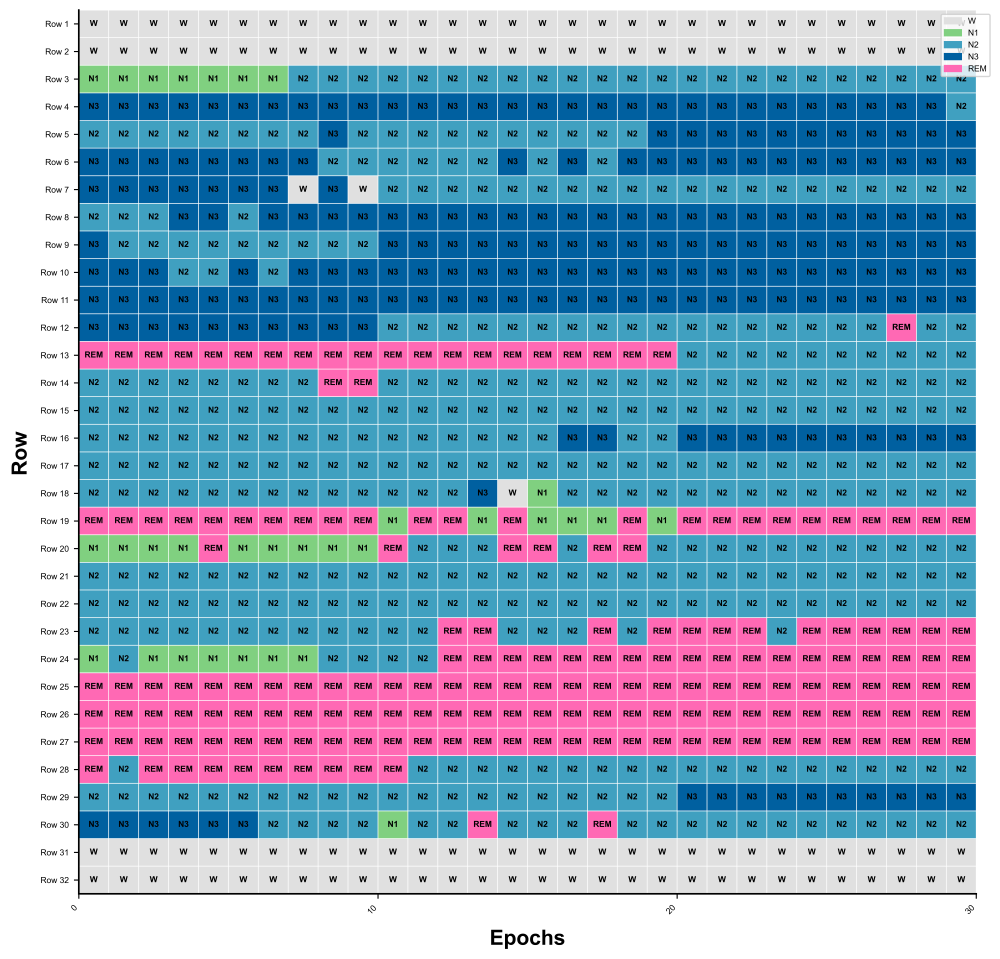}
\caption{Train Transformer}
\label{fig:vis_d}
\end{subfigure}

\caption{
Visualization of sleep stage transitions for subject \texttt{SC4181E0}. Different colors represent distinct sleep stages. (a) Ground truth hypnogram. (b) Raw Epoch Encoder predictions showing fragmented transitions. (c) Random Transformer improves temporal consistency. (d) Fully trained Transformer results.
}
\label{fig:sc4181e0_vis}

\end{figure}

We present a qualitative visualization of whole-night sleep stage transitions for subject \texttt{SC4181E0}. As shown in Figure~\ref{fig:sc4181e0_vis}, the Epoch Encoder (Fig.~\ref{fig:vis_a}) generates highly fragmented predictions with frequent unrealistic transitions, reflecting the absence of sequence context. Applying the Random Transformer (Fig.~\ref{fig:vis_b}) substantially stabilizes these predictions by smoothing transitions within local windows, resulting in a hypnogram that more closely aligns with the Ground Truth (Fig.~\ref{fig:vis_c}). The trained Transformer (Fig.~\ref{fig:vis_d}) produces sequences with similar smoothness, yet offers no clear advantage compared with the Random Transformer. This outcome underscores that most of the sequence-level correction arises from the structural prior embedded in random self-attention rather than from supervised optimization.

   

   

\section{Discussion, Limitation, Conclusion and Future Work}
\label{sec:Limitations, Conclusion and Future Work}




Our findings reveal a striking and counter-intuitive phenomenon: the effectiveness of Transformers in sleep staging may arise far less from learned long-range dependency modeling than previously assumed. Through RAPK, we show that even randomly initialized Transformers naturally behave as adaptive sequence smoothers, suggesting that much of the performance gain in modern sleep staging originates from architectural inductive bias itself. This challenges the prevailing paradigm that increasingly complex sequence learning is the primary driver of progress in physiological AI. 

However, smoothing-based mechanisms may struggle to capture rare, abrupt, or pathological transitions, highlighting the need for future models that better balance adaptive smoothing and sensitivity to clinically important sparse events. Another important direction is improving encoder training to achieve stronger representation learning for downstream sleep staging.

These findings further suggest that future sleep foundation models should prioritize stronger epoch-level representations while relying on simpler and more structure-driven sequence modules. RAPK also indicates that lightweight, shallow multi-head smoothing architectures are naturally suited for scalable mobile and edge deployment in real-world sleep monitoring.

\bibliography{example_paper}

\begin{thebibliography}{47}
\providecommand{\natexlab}[1]{#1}
\providecommand{\url}[1]{\texttt{#1}}
\expandafter\ifx\csname urlstyle\endcsname\relax
  \providecommand{\doi}[1]{doi: #1}\else
  \providecommand{\doi}{doi: \begingroup \urlstyle{rm}\Url}\fi

\bibitem[Luyster et~al.(2012)Luyster, Strollo, Zee, and Walsh]{10.5665/sleep.1846}
Faith~S. Luyster, Jr. Strollo, Patrick~J., Phyllis~C. Zee, and James~K. Walsh.
\newblock Sleep: A health imperative.
\newblock \emph{Sleep}, 35\penalty0 (6):\penalty0 727--734, 06 2012.
\newblock ISSN 0161-8105.
\newblock \doi{10.5665/sleep.1846}.
\newblock URL \url{https://doi.org/10.5665/sleep.1846}.

\bibitem[Phan and Mikkelsen(2022)]{phan2022automatic}
Huy Phan and Kaare Mikkelsen.
\newblock Automatic sleep staging of eeg signals: recent development, challenges, and future directions.
\newblock \emph{Physiological measurement}, 43\penalty0 (4):\penalty0 04TR01, 2022.

\bibitem[Sun et~al.(2017)Sun, Jia, Goparaju, Huang, Sourina, Bianchi, and Westover]{10.1093/sleep/zsx139}
Haoqi Sun, Jian Jia, Balaji Goparaju, Guang-Bin Huang, Olga Sourina, Matt~Travis Bianchi, and M~Brandon Westover.
\newblock Large-scale automated sleep staging.
\newblock \emph{Sleep}, 40\penalty0 (10):\penalty0 zsx139, 09 2017.
\newblock ISSN 0161-8105.
\newblock \doi{10.1093/sleep/zsx139}.
\newblock URL \url{https://doi.org/10.1093/sleep/zsx139}.

\bibitem[Liu et~al.(2025)Liu, Zhang, Luo, Wei, Sun, Deng, Wei, and Chen]{liu2025sleepmodulationchallengetransitioning}
Guisong Liu, Jiansong Zhang, Yinpei Luo, Guoliang Wei, Shuqing Sun, Shiyang Deng, Pengfei Wei, and Nanxi Chen.
\newblock Sleep modulation: The challenge of transitioning from open loop to closed loop, 2025.
\newblock URL \url{https://arxiv.org/abs/2512.03784}.

\bibitem[Krugliakova et~al.(2026)Krugliakova, Breuer, Adelh\"{o}fer, Alonso, Besedovsky, Murphy, Peters, Raczek, Rasch, Salvesen, Snipes, Schoch, Schreiner, Wassing, Bergmann, and Dresler]{doi:10.1152/physrev.00007.2025}
Elena Krugliakova, Friederike Breuer, Nico Adelh\"{o}fer, Alejandra Alonso, Luciana Besedovsky, Keith Murphy, Emma Peters, Karolina Raczek, Bj\"{o}rn Rasch, Leila Salvesen, Sophia Snipes, Sarah Schoch, Thomas Schreiner, Rick Wassing, Til~Ole Bergmann, and Martin Dresler.
\newblock Hacking the functions of sleep: noninvasive approaches to stimulate sleep neurophysiology.
\newblock \emph{Physiological Reviews}, 106\penalty0 (2):\penalty0 675--749, 2026.
\newblock \doi{10.1152/physrev.00007.2025}.
\newblock URL \url{https://doi.org/10.1152/physrev.00007.2025}.
\newblock PMID: 41263765.

\bibitem[Phan et~al.(2023)Phan, Lorenzen, Heremans, Chén, Tran, Koch, Mertins, Baumert, Mikkelsen, and De~Vos]{10210638}
Huy Phan, Kristian~P. Lorenzen, Elisabeth Heremans, Oliver~Y. Chén, Minh~C. Tran, Philipp Koch, Alfred Mertins, Mathias Baumert, Kaare~B. Mikkelsen, and Maarten De~Vos.
\newblock L-seqsleepnet: Whole-cycle long sequence modeling for automatic sleep staging.
\newblock \emph{IEEE Journal of Biomedical and Health Informatics}, 27\penalty0 (10):\penalty0 4748--4757, 2023.
\newblock \doi{10.1109/JBHI.2023.3303197}.

\bibitem[Colten and Altevogt(2006)]{book}
H.R. Colten and Bruce Altevogt.
\newblock \emph{Sleep disorders and sleep deprivation: An unmet public health problem}.
\newblock National Academies Press, 10 2006.
\newblock ISBN 9780309101110.
\newblock \doi{10.17226/11617}.

\bibitem[Berry et~al.(2017)Berry, Brooks, Gamaldo, Harding, Lloyd, Quan, Troester, and Vaughn]{doi:10.5664/jcsm.6576}
Richard~B. Berry, Rita Brooks, Charlene Gamaldo, Susan~M. Harding, Robin~M. Lloyd, Stuart~F. Quan, Matthew~T. Troester, and Bradley~V. Vaughn.
\newblock Aasm scoring manual updates for 2017 (version 2.4).
\newblock \emph{Journal of Clinical Sleep Medicine}, 13\penalty0 (05):\penalty0 665--666, 2017.
\newblock \doi{10.5664/jcsm.6576}.
\newblock URL \url{https://jcsm.aasm.org/doi/abs/10.5664/jcsm.6576}.

\bibitem[Liao et~al.(2022)Liao, Zhang, Zhang, Wang, and Xie]{Liao_2022}
Yiqiao Liao, Chao Zhang, Milin Zhang, Zhihua Wang, and Xiang Xie.
\newblock Lightsleepnet: Design of a personalized portable sleep staging system based on single-channel eeg.
\newblock \emph{IEEE Transactions on Circuits and Systems II: Express Briefs}, 69\penalty0 (1):\penalty0 224–228, January 2022.
\newblock ISSN 1558-3791.
\newblock \doi{10.1109/tcsii.2021.3086981}.
\newblock URL \url{http://dx.doi.org/10.1109/TCSII.2021.3086981}.

\bibitem[Liu et~al.(2023)Liu, Wei, Sun, Mao, Zhang, Zhao, Tian, Wang, and Chen]{10.3389/fnins.2023.1218072}
Guisong Liu, Guoliang Wei, Shuqing Sun, Dandan Mao, Jiansong Zhang, Dechun Zhao, Xuelong Tian, Xing Wang, and Nanxi Chen.
\newblock Micro sleepnet: efficient deep learning model for mobile terminal real-time sleep staging.
\newblock \emph{Frontiers in Neuroscience}, Volume 17 - 2023, 2023.
\newblock ISSN 1662-453X.
\newblock \doi{10.3389/fnins.2023.1218072}.
\newblock URL \url{https://www.frontiersin.org/journals/neuroscience/articles/10.3389/fnins.2023.1218072}.

\bibitem[Supratak et~al.(2017)Supratak, Dong, Wu, and Guo]{7961240}
Akara Supratak, Hao Dong, Chao Wu, and Yike Guo.
\newblock Deepsleepnet: A model for automatic sleep stage scoring based on raw single-channel eeg.
\newblock \emph{IEEE Transactions on Neural Systems and Rehabilitation Engineering}, 25\penalty0 (11):\penalty0 1998--2008, 2017.
\newblock \doi{10.1109/TNSRE.2017.2721116}.

\bibitem[Supratak and Guo(2020)]{9176741}
Akara Supratak and Yike Guo.
\newblock Tinysleepnet: An efficient deep learning model for sleep stage scoring based on raw single-channel eeg.
\newblock In \emph{2020 42nd Annual International Conference of the IEEE Engineering in Medicine \& Biology Society (EMBC)}, pages 641--644, 2020.
\newblock \doi{10.1109/EMBC44109.2020.9176741}.

\bibitem[Yu et~al.(2025)Yu, Hu, He, Wu, Gu, Yu, Li, Wang, and Xiao]{10965348}
Tianyou Yu, Xinxin Hu, Yanbin He, Wei Wu, Zhenghui Gu, Zhuliang Yu, Yuanqing Li, Fei Wang, and Jun Xiao.
\newblock Multi-view self-supervised learning enhances automatic sleep staging from eeg signals.
\newblock \emph{IEEE Transactions on Biomedical Engineering}, 72\penalty0 (10):\penalty0 3056--3070, 2025.
\newblock \doi{10.1109/TBME.2025.3561228}.

\bibitem[Thapa et~al.(2025)Thapa, Kj{\ae}r, He, Covert, Moore, Hanif, Ganjoo, Westover, Jennum, Brink-Kj{\ae}r, et~al.]{thapa2025multimodal}
Rahul Thapa, Magnus~Ruud Kj{\ae}r, Bryan He, Ian Covert, Hyatt Moore, Umaer Hanif, Gauri Ganjoo, Brandon~M Westover, Poul Jennum, Andreas Brink-Kj{\ae}r, et~al.
\newblock A multimodal sleep foundation model developed with 500k hours of sleep recordings for disease predictions.
\newblock \emph{medRxiv}, pages 2025--02, 2025.

\bibitem[Eldele et~al.(2021)Eldele, Chen, Liu, Wu, Kwoh, Li, and Guan]{9417097}
Emadeldeen Eldele, Zhenghua Chen, Chengyu Liu, Min Wu, Chee-Keong Kwoh, Xiaoli Li, and Cuntai Guan.
\newblock An attention-based deep learning approach for sleep stage classification with single-channel eeg.
\newblock \emph{IEEE Transactions on Neural Systems and Rehabilitation Engineering}, 29:\penalty0 809--818, 2021.
\newblock \doi{10.1109/TNSRE.2021.3076234}.

\bibitem[Phan et~al.(2022)Phan, Mikkelsen, Chén, Koch, Mertins, and De~Vos]{9697331}
Huy Phan, Kaare Mikkelsen, Oliver~Y. Chén, Philipp Koch, Alfred Mertins, and Maarten De~Vos.
\newblock Sleeptransformer: Automatic sleep staging with interpretability and uncertainty quantification.
\newblock \emph{IEEE Transactions on Biomedical Engineering}, 69\penalty0 (8):\penalty0 2456--2467, 2022.
\newblock \doi{10.1109/TBME.2022.3147187}.

\bibitem[Deng et~al.(2024)Deng, Niu, Rao, Luo, Zhang, Xie, Yu, Liu, Zhang, Zhao, Pan, Li, Deng, Guo, Zhang, Li, and Jiang]{deng2024lpsgm}
Guifeng Deng, Mengfan Niu, Shuying Rao, Yuxi Luo, Jianjia Zhang, Junyi Xie, Zhenghe Yu, Wenjuan Liu, Junhang Zhang, Sha Zhao, Gang Pan, Xiaojing Li, Wei Deng, Wanjun Guo, Yaoyun Zhang, Tao Li, and Haiteng Jiang.
\newblock A unified flexible large psg model for sleep staging and brain disorder diagnosis.
\newblock \emph{medRxiv}, 2024.
\newblock \doi{10.1101/2024.12.11.24318815}.
\newblock URL \url{https://www.medrxiv.org/content/early/2025/11/27/2024.12.11.24318815}.

\bibitem[Wang et~al.(2024)Wang, Zhao, Jiang, Zhou, Yu, Li, Li, and Pan]{10595067}
Jiquan Wang, Sha Zhao, Haiteng Jiang, Yangxuan Zhou, Zhenghe Yu, Tao Li, Shijian Li, and Gang Pan.
\newblock Caresleepnet: A hybrid deep learning network for automatic sleep staging.
\newblock \emph{IEEE Journal of Biomedical and Health Informatics}, 28\penalty0 (12):\penalty0 7392--7405, 2024.
\newblock \doi{10.1109/JBHI.2024.3426939}.

\bibitem[Chin et~al.(2025)Chin, Yew, Wu, Liang, Chan, Zain, Samdin, and Goh]{2025sleepdifformer}
Benjamin Wei~Hao Chin, Yuin~Torng Yew, Haocheng Wu, Lanxin Liang, Chow~Khuen Chan, Norita~Mohd Zain, Siti~Balqis Samdin, and Sim~Kuan Goh.
\newblock Sleepdifformer: Sleep stage classification via multivariate differential transformer.
\newblock \emph{arXiv preprint arXiv:2508.15215}, 2025.

\bibitem[Guo et~al.(2024)Guo, Nowakowski, and Dai]{guo_flexsleeptransformer_2024}
Yanchen Guo, Maciej Nowakowski, and Weiying Dai.
\newblock {FlexSleepTransformer}: a transformer-based sleep staging model with flexible input channel configurations.
\newblock 14\penalty0 (1):\penalty0 26312, 2024.
\newblock ISSN 2045-2322.
\newblock \doi{10.1038/s41598-024-76197-0}.
\newblock URL \url{https://doi.org/10.1038/s41598-024-76197-0}.

\bibitem[Lee et~al.(2025)Lee, Choi, Lee, Jeong, Hong, Shin, and Kim]{lee_explainable_2025}
Hyojin Lee, You~Rim Choi, Hyun~Kyung Lee, Jaemin Jeong, Joopyo Hong, Hyun-Woo Shin, and Hyung-Sin Kim.
\newblock Explainable vision transformer for automatic visual sleep staging on multimodal {PSG} signals.
\newblock 8\penalty0 (1):\penalty0 55, 2025.
\newblock ISSN 2398-6352.
\newblock \doi{10.1038/s41746-024-01378-0}.
\newblock URL \url{https://doi.org/10.1038/s41746-024-01378-0}.

\bibitem[Coon and Ogg(2024)]{10782964}
William~G. Coon and Mattson Ogg.
\newblock Laying the foundation: Modern transformers for gold-standard sleep analysis and beyond.
\newblock In \emph{2024 46th Annual International Conference of the IEEE Engineering in Medicine and Biology Society (EMBC)}, pages 1--7, 2024.
\newblock \doi{10.1109/EMBC53108.2024.10782964}.

\bibitem[Ciudad et~al.(2024)Ciudad, Mørup, Kornum, and Zahid]{10782771}
Javier~García Ciudad, Morten Mørup, Birgitte~Rahbek Kornum, and Alexander~Neergaard Zahid.
\newblock Evaluating the influence of temporal context on automatic mouse sleep staging through the application of human models.
\newblock In \emph{2024 46th Annual International Conference of the IEEE Engineering in Medicine and Biology Society (EMBC)}, pages 1--4, 2024.
\newblock \doi{10.1109/EMBC53108.2024.10782771}.

\bibitem[Pradeepkumar et~al.(2024)Pradeepkumar, Anandakumar, Kugathasan, Suntharalingham, Kappel, De~Silva, and Edussooriya]{10623416}
Jathurshan Pradeepkumar, Mithunjha Anandakumar, Vinith Kugathasan, Dhinesh Suntharalingham, Simon~L. Kappel, Anjula~C. De~Silva, and Chamira U.~S. Edussooriya.
\newblock Toward interpretable sleep stage classification using cross-modal transformers.
\newblock \emph{IEEE Transactions on Neural Systems and Rehabilitation Engineering}, 32:\penalty0 2893--2904, 2024.
\newblock \doi{10.1109/TNSRE.2024.3438610}.

\bibitem[Vaswani et~al.(2017)Vaswani, Shazeer, Parmar, Uszkoreit, Jones, Gomez, Kaiser, and Polosukhin]{10.5555/3295222.3295349}
Ashish Vaswani, Noam Shazeer, Niki Parmar, Jakob Uszkoreit, Llion Jones, Aidan~N. Gomez, \L{}ukasz Kaiser, and Illia Polosukhin.
\newblock Attention is all you need.
\newblock In \emph{Proceedings of the 31st International Conference on Neural Information Processing Systems}, NIPS'17, page 6000–6010, Red Hook, NY, USA, 2017. Curran Associates Inc.
\newblock ISBN 9781510860964.

\bibitem[Phan et~al.(2019)Phan, Andreotti, Cooray, Chén, and De~Vos]{8631195}
Huy Phan, Fernando Andreotti, Navin Cooray, Oliver~Y. Chén, and Maarten De~Vos.
\newblock Seqsleepnet: End-to-end hierarchical recurrent neural network for sequence-to-sequence automatic sleep staging.
\newblock \emph{IEEE Transactions on Neural Systems and Rehabilitation Engineering}, 27\penalty0 (3):\penalty0 400--410, 2019.
\newblock \doi{10.1109/TNSRE.2019.2896659}.

\bibitem[Lukoševičius and Jaeger()]{lukosevicius_reservoir_2009}
Mantas Lukoševičius and Herbert Jaeger.
\newblock Reservoir computing approaches to recurrent neural network training.
\newblock 3\penalty0 (3):\penalty0 127--149.
\newblock ISSN 1574-0137.
\newblock \doi{https://doi.org/10.1016/j.cosrev.2009.03.005}.
\newblock URL \url{https://www.sciencedirect.com/science/article/pii/S1574013709000173}.

\bibitem[Seo et~al.(2020)Seo, Back, Lee, Park, Kim, and Lee]{seo2020intra}
Hogeon Seo, Seunghyeok Back, Seongju Lee, Deokhwan Park, Tae Kim, and Kyoobin Lee.
\newblock Intra-and inter-epoch temporal context network (iitnet) using sub-epoch features for automatic sleep scoring on raw single-channel eeg.
\newblock \emph{Biomedical signal processing and control}, 61:\penalty0 102037, 2020.

\bibitem[Rahimi and Recht(2007)]{10.5555/2981562.2981710}
Ali Rahimi and Benjamin Recht.
\newblock Random features for large-scale kernel machines.
\newblock In \emph{Proceedings of the 21st International Conference on Neural Information Processing Systems}, NIPS'07, page 1177–1184, Red Hook, NY, USA, 2007. Curran Associates Inc.
\newblock ISBN 9781605603520.

\bibitem[Gallicchio and Micheli(2020)]{gallicchio2020deepechostatenetwork}
Claudio Gallicchio and Alessio Micheli.
\newblock Deep echo state network (deepesn): A brief survey, 2020.
\newblock URL \url{https://arxiv.org/abs/1712.04323}.

\bibitem[Lempitsky et~al.(2018)Lempitsky, Vedaldi, and Ulyanov]{8579082}
Victor Lempitsky, Andrea Vedaldi, and Dmitry Ulyanov.
\newblock Deep image prior.
\newblock In \emph{2018 IEEE/CVF Conference on Computer Vision and Pattern Recognition}, pages 9446--9454, 2018.
\newblock \doi{10.1109/CVPR.2018.00984}.

\bibitem[Choromanski et~al.(2020)Choromanski, Likhosherstov, Dohan, Song, Gane, Sarlos, Hawkins, Davis, Mohiuddin, Kaiser, et~al.]{choromanski2020rethinking}
Krzysztof Choromanski, Valerii Likhosherstov, David Dohan, Xingyou Song, Andreea Gane, Tamas Sarlos, Peter Hawkins, Jared Davis, Afroz Mohiuddin, Lukasz Kaiser, et~al.
\newblock Rethinking attention with performers.
\newblock \emph{arXiv preprint arXiv:2009.14794}, 2020.

\bibitem[Peng et~al.(2021)Peng, Pappas, Yogatama, Schwartz, Smith, and Kong]{peng2021randomfeatureattention}
Hao Peng, Nikolaos Pappas, Dani Yogatama, Roy Schwartz, Noah~A. Smith, and Lingpeng Kong.
\newblock Random feature attention, 2021.
\newblock URL \url{https://arxiv.org/abs/2103.02143}.

\bibitem[Tsai et~al.(2019)Tsai, Bai, Yamada, Morency, and Salakhutdinov]{tsai-etal-2019-transformer}
Yao-Hung~Hubert Tsai, Shaojie Bai, Makoto Yamada, Louis-Philippe Morency, and Ruslan Salakhutdinov.
\newblock Transformer dissection: An unified understanding for transformer{'}s attention via the lens of kernel.
\newblock In Kentaro Inui, Jing Jiang, Vincent Ng, and Xiaojun Wan, editors, \emph{Proceedings of the 2019 Conference on Empirical Methods in Natural Language Processing and the 9th International Joint Conference on Natural Language Processing (EMNLP-IJCNLP)}, pages 4344--4353, Hong Kong, China, November 2019. Association for Computational Linguistics.
\newblock \doi{10.18653/v1/D19-1443}.
\newblock URL \url{https://aclanthology.org/D19-1443/}.

\bibitem[Park and Kim(2022)]{park2022visiontransformerswork}
Namuk Park and Songkuk Kim.
\newblock How do vision transformers work?, 2022.
\newblock URL \url{https://arxiv.org/abs/2202.06709}.

\bibitem[Wang et~al.(2022)Wang, Zheng, Chen, and Wang]{wang2022antioversmoothingdeepvisiontransformers}
Peihao Wang, Wenqing Zheng, Tianlong Chen, and Zhangyang Wang.
\newblock Anti-oversmoothing in deep vision transformers via the fourier domain analysis: From theory to practice, 2022.
\newblock URL \url{https://arxiv.org/abs/2203.05962}.

\bibitem[Jacot et~al.(2021)Jacot, Gabriel, and Hongler]{10.1145/3406325.3465355}
Arthur Jacot, Franck Gabriel, and Cl\'{e}ment Hongler.
\newblock Neural tangent kernel: convergence and generalization in neural networks (invited paper).
\newblock In \emph{Proceedings of the 53rd Annual ACM SIGACT Symposium on Theory of Computing}, STOC 2021, page~6, New York, NY, USA, 2021. Association for Computing Machinery.
\newblock ISBN 9781450380539.
\newblock \doi{10.1145/3406325.3465355}.
\newblock URL \url{https://doi.org/10.1145/3406325.3465355}.

\bibitem[Hron et~al.(2020)Hron, Bahri, Sohl-Dickstein, and Novak]{hron2020infiniteattentionnngpntk}
Jiri Hron, Yasaman Bahri, Jascha Sohl-Dickstein, and Roman Novak.
\newblock Infinite attention: Nngp and ntk for deep attention networks, 2020.
\newblock URL \url{https://arxiv.org/abs/2006.10540}.

\bibitem[Goldberger et~al.(2000)Goldberger, Amaral, Glass, Hausdorff, Ivanov, Mark, Mietus, Moody, Peng, and Stanley]{doi:10.1161/01.CIR.101.23.e215}
Ary~L. Goldberger, Luis A.~N. Amaral, Leon Glass, Jeffrey~M. Hausdorff, Plamen~Ch. Ivanov, Roger~G. Mark, Joseph~E. Mietus, George~B. Moody, Chung-Kang Peng, and H.~Eugene Stanley.
\newblock Physiobank, physiotoolkit, and physionet.
\newblock \emph{Circulation}, 101\penalty0 (23):\penalty0 e215--e220, 2000.
\newblock \doi{10.1161/01.CIR.101.23.e215}.
\newblock URL \url{https://www.ahajournals.org/doi/abs/10.1161/01.CIR.101.23.e215}.

\bibitem[Quan et~al.(1997)Quan, Howard, Iber, Kiley, Nieto, O'Connor, Rapoport, Redline, Robbins, Samet, and Wahl]{10.1093/sleep/20.12.1077}
Stuart~F. Quan, Barbara~V. Howard, Conrad Iber, James~P. Kiley, F.~Javier Nieto, George~T. O'Connor, David~M. Rapoport, Susan Redline, John Robbins, Jonathan~M. Samet, and ‡Patricia~W. Wahl.
\newblock The sleep heart health study: Design, rationale, and methods.
\newblock \emph{Sleep}, 20\penalty0 (12):\penalty0 1077--1085, 12 1997.
\newblock ISSN 0161-8105.
\newblock \doi{10.1093/sleep/20.12.1077}.
\newblock URL \url{https://doi.org/10.1093/sleep/20.12.1077}.

\bibitem[Zhou et~al.(2022)Zhou, Wei, Wang, Shen, Xie, Yuille, and Kong]{zhou2021ibot}
Jinghao Zhou, Chen Wei, Huiyu Wang, Wei Shen, Cihang Xie, Alan Yuille, and Tao Kong.
\newblock ibot: Image bert pre-training with online tokenizer.
\newblock \emph{International Conference on Learning Representations (ICLR)}, 2022.

\bibitem[Hochreiter and Schmidhuber(1997)]{6795963}
Sepp Hochreiter and Jürgen Schmidhuber.
\newblock Long short-term memory.
\newblock \emph{Neural Computation}, 9\penalty0 (8):\penalty0 1735--1780, 1997.
\newblock \doi{10.1162/neco.1997.9.8.1735}.

\bibitem[Chung et~al.(2014)Chung, Gulcehre, Cho, and Bengio]{chung2014empiricalevaluationgatedrecurrent}
Junyoung Chung, Caglar Gulcehre, KyungHyun Cho, and Yoshua Bengio.
\newblock Empirical evaluation of gated recurrent neural networks on sequence modeling, 2014.
\newblock URL \url{https://arxiv.org/abs/1412.3555}.

\bibitem[Lea et~al.(2017)Lea, Flynn, Vidal, Reiter, and Hager]{8099596}
Colin Lea, Michael~D. Flynn, René Vidal, Austin Reiter, and Gregory~D. Hager.
\newblock Temporal convolutional networks for action segmentation and detection.
\newblock In \emph{2017 IEEE Conference on Computer Vision and Pattern Recognition (CVPR)}, pages 1003--1012, 2017.
\newblock \doi{10.1109/CVPR.2017.113}.

\bibitem[Gu and Dao(2024)]{gu2024mambalineartimesequencemodeling}
Albert Gu and Tri Dao.
\newblock Mamba: Linear-time sequence modeling with selective state spaces, 2024.
\newblock URL \url{https://arxiv.org/abs/2312.00752}.

\bibitem[Grieger et~al.(2025)Grieger, Raskob, Mehrkanoon, and Bialonski]{grieger2025anysleepchannelagnosticdeeplearning}
Niklas Grieger, Jannik Raskob, Siamak Mehrkanoon, and Stephan Bialonski.
\newblock Anysleep: a channel-agnostic deep learning system for high-resolution sleep staging in multi-center cohorts, 2025.
\newblock URL \url{https://arxiv.org/abs/2512.14461}.

\bibitem[Fonseca et~al.(2017)Fonseca, den Teuling, Long, and Aarts]{7446264}
Pedro Fonseca, Niek den Teuling, Xi~Long, and Ronald~M. Aarts.
\newblock Cardiorespiratory sleep stage detection using conditional random fields.
\newblock \emph{IEEE Journal of Biomedical and Health Informatics}, 21\penalty0 (4):\penalty0 956--966, 2017.
\newblock \doi{10.1109/JBHI.2016.2550104}.

\end{thebibliography}
\bibliographystyle{unsrtnat}

\newpage
\appendix
\onecolumn
\appendix
\appendix
\section{Framework and Model Details}
\label{app:Model Details}  
\subsection{Overview of the Framework}
We adopt a two-stage automatic sleep staging framework. Unlike prior methods, this study focuses on the effect of the intrinsic random initialization bias of the Transformer architecture on temporal modeling:
\begin{itemize}
    \item \textbf{Stage 1:} A Transformer-based \emph{Epoch Encoder} extracts local features from single 30-second EEG segments.
    \item \textbf{Stage 2:} A randomly initialized and \emph{untrained} Temporal Transformer models sequences across epochs.
\end{itemize}
This design allows independent investigation of the natural temporal smoothing and contextual dependencies modeled by the architecture itself, without parameter optimization.

\subsection{Epoch Encoder}
The Epoch Encoder extracts local time-frequency features from a 30-second EEG segment and maps them to a latent representation. The input is $X^{(i)}\in\mathbb{R}^{1\times T}$, $T=3000$, with a sampling rate of 100 Hz.

\subsubsection{Patch Embedding and Linear Projection}
The input sequence is divided into fixed-length patches $P=100$:
\[
X^{(i)}=[x_1^{(i)},\dots,x_{N_p}^{(i)}],\quad x_j^{(i)}\in\mathbb{R}^P, \quad N_p=\lceil T/P\rceil.
\]
Each patch is linearly projected via a 1D convolution:
\[
h_j^{(i)}=W_p x_j^{(i)}+b_p, \quad W_p\in\mathbb{R}^{d\times P},\ b_p\in\mathbb{R}^{d}, \quad d=128,
\]
yielding a patch sequence:
\[
H^{(i)}=[h_1^{(i)},\dots,h_{N_p}^{(i)}]\in\mathbb{R}^{N_p\times d}.
\]

\subsubsection{Positional Encoding and \texorpdfstring{[CLS]}{CLS} Token}
A learnable [CLS] token aggregates epoch-level semantics:
\[
H_\text{in}^{(i)}=[h_\text{cls}; H^{(i)}]\in\mathbb{R}^{(N_p+1)\times d}.
\]
Adding learnable positional embeddings $P_\text{patch}\in\mathbb{R}^{(N_p+1)\times d}$:
\[
\tilde{H}_\text{in}^{(i)}=H_\text{in}^{(i)}+P_\text{patch}.
\]

\subsubsection{Transformer Encoding}
The sequence is input to a Transformer encoder:
\[
\tilde{H}_\text{out}^{(i)} = \text{TransformerEncoder}(\tilde{H}_\text{in}^{(i)};\theta_\text{epoch}).
\]

\subsubsection{Epoch-level Feature Vector}
The [CLS] output serves as the epoch-level representation:
\[
z^{(i)} = \tilde{H}_\text{out}^{(i)}[0] \in \mathbb{R}^{d}.
\]

\subsection{Temporal Transformer Encoder}
After encoding, consecutive epochs form a latent sequence:
\[
Z = \{ z^{(1)}, \dots, z^{(N)} \}, \quad z^{(i)}\in\mathbb{R}^{d}, \quad N=10.
\]

\subsubsection{Temporal Positional Encoding}
A learnable temporal positional embedding $P_\text{temp}\in\mathbb{R}^{N\times d}$ is added:
\[
\tilde{Z} = Z + P_\text{temp}.
\]

\subsubsection{Temporal Modeling Layer}
The sequence with temporal embeddings is input to a temporal Transformer:
\[
\tilde{H}_\text{out} = \text{TransformerEncoder}(\tilde{Z};\theta_\text{temp}),
\]
enabling global self-attention across all epochs.

\subsection{Sleep Stage Classification}
For each epoch $i$, the context-enhanced output $\tilde{h}_\text{out}^{(i)} \in \mathbb{R}^d$ is classified using a linear layer followed by softmax:
\[
\hat{y}^{(i)} = \text{Softmax}(W_c \tilde{h}_\text{out}^{(i)} + b_c), \quad W_c \in \mathbb{R}^{K\times d}, \ b_c \in \mathbb{R}^K,
\]
where $K$ denotes the number of sleep stages (e.g., W, N1, N2, N3, REM). The sequence output is
\[
\hat{Y} = [\hat{y}^{(1)}, \dots, \hat{y}^{(N)}] \in \mathbb{R}^{N\times K}.
\]

\section{Expectation Derivation of RAPK (Initialization-Independent)}
\label{app:rapk_derivation}  
\subsection{Notation, Assumptions, and Objective}
Let the input matrix be 
\[
X=[x_1^\top; \dots; x_T^\top] \in \mathbb{R}^{T \times d}, \quad x_p \in \mathbb{R}^d.
\]
Random linear layers $W_Q, W_K, W_V \in \mathbb{R}^{d \times d_k}$ satisfy:
\begin{itemize}
    \item i.i.d.\ elements, zero mean: $E[W_Q] = E[W_K] = E[W_V] = 0$
    \item finite variance: $\text{Var}[(W_Q)_{ab}] = \sigma_Q^2, \text{Var}[(W_K)_{ab}] = \sigma_K^2, \text{Var}[(W_V)_{ab}] = \sigma_V^2$
    \item mutual independence: $W_Q \perp W_K \perp W_V$
\end{itemize}

Define
\[
Q = X W_Q, \quad K = X W_K, \quad V = X W_V.
\]
For the $i$-th row:
\[
q_i = x_i^\top W_Q, \quad k_p = x_p^\top W_K, \quad v_p = x_p^\top W_V.
\]
Scaled attention scores:
\[
s_{ip} = \frac{q_i \cdot k_p}{\sqrt{d_k}}, \quad 
A = \text{Softmax}\left(\frac{QK^\top}{\sqrt{d_k}}\right), \quad 
a_{ip} = A_{ip}.
\]
Output matrix:
\[
O = AV, \quad o_i = \sum_{p=1}^T a_{ip} v_p.
\]
Random Attention Prior Kernel (RAPK):
\[
K_\text{RAP} = O O^\top.
\]

The goal is to derive
\[
E[K_\text{RAP}]_{ij} \approx C_0 + C_1 (x_i^\top x_j).
\]

\subsection{Second Moment of Value Vectors}
\[
v_p^\top v_q = (x_p^\top W_V)(W_V^\top x_q) = x_p^\top W_V W_V^\top x_q.
\]
Taking expectation over $W_V$:
\[
E_{W_V}[v_p^\top v_q] = x_p^\top \, E[W_V W_V^\top] \, x_q.
\]
Since columns of $W_V$ are i.i.d.\ with zero mean and variance $\sigma_V^2$:
\[
E[W_V W_V^\top] = \sum_{r=1}^{d_k} E[w_r w_r^\top] = d_k \sigma_V^2 I_d.
\]
Hence,
\begin{equation}
E_{W_V}[v_p^\top v_q] = d_k \sigma_V^2 (x_p^\top x_q). \label{eq:V_second_moment}
\end{equation}

Substitute into $K_\text{RAP}$:
\[
K_\text{RAP} = \sum_{p,q=1}^T a_{ip} a_{jq} (v_p^\top v_q), \quad 
E[K_{ij}] = \sum_{p,q} E_{W_Q,W_K}[a_{ip} a_{jq}] \cdot E_{W_V}[v_p^\top v_q].
\]
Using \eqref{eq:V_second_moment}:
\begin{equation}
E[K_{ij}] = d_k \sigma_V^2 \sum_{p,q} E_{W_Q,W_K}[a_{ip} a_{jq}] (x_p^\top x_q). \label{eq:K_RAP_E}
\end{equation}

\subsection{Linearization of Softmax}
Define $S_i = \sum_{t=1}^T s_{it}$. Using first-order Taylor expansion:
\[
e^{s_{ip}} = 1 + s_{ip} + \frac{1}{2}s_{ip}^2 + O(s^3).
\]
Numerator:
\[
\text{Num} = 1 + s_{ip} + O(s^2), \quad 
\text{Den} = \sum_{t=1}^T (1 + s_{it} + O(s^2)) = T + S_i + O(\sum s_{it}^2).
\]
Reciprocal expansion:
\[
\frac{1}{T + S_i + \epsilon} \approx \frac{1}{T}\left(1 - \frac{S_i}{T} - \frac{\epsilon}{T}\right) + O(s^2).
\]
Thus,
\[
a_{ip} \approx \frac{1}{T} + \frac{1}{T} (s_{ip} - \bar{s}_i) + O(s^2), \quad \bar{s}_i = \frac{S_i}{T}.
\]
For the product:
\[
a_{ip} a_{jq} \approx \frac{1}{T^2} + \frac{1}{T^2} [(s_{ip} - \bar{s}_i) + (s_{jq} - \bar{s}_j)] + \frac{1}{T^2} (s_{ip} - \bar{s}_i)(s_{jq} - \bar{s}_j) + O(s^3).
\]
Expectation:
\begin{equation}
E[a_{ip} a_{jq}] \approx \frac{1}{T^2} + \frac{1}{T^2} E[(s_{ip} - \bar{s}_i)(s_{jq} - \bar{s}_j)]. \label{eq:a_ip_a_jq}
\end{equation}

\subsection{Computation of $E[s_{ip}s_{jq}]$}
Write
\[
q_i = \sum_{r=1}^{d_k} (x_i^\top w_r^Q) e_r, \quad k_p = \sum_{r=1}^{d_k} (x_p^\top w_r^K) e_r,
\]
so that
\[
s_{ip} = \frac{1}{\sqrt{d_k}} \sum_{r=1}^{d_k} u_{i,r}^Q u_{p,r}^K, \quad u_{i,r}^Q = x_i^\top w_r^Q, \ u_{p,r}^K = x_p^\top w_r^K.
\]
Then
\[
s_{ip} s_{jq} = \frac{1}{d_k} \sum_{r,r'=1}^{d_k} u_{i,r}^Q u_{p,r}^K u_{j,r'}^Q u_{q,r'}^K.
\]
Independence implies only terms with $r=r'$ remain:
\[
E[s_{ip} s_{jq}] = \frac{1}{d_k} \sum_{r=1}^{d_k} E[u_{i,r}^Q u_{j,r}^Q] \, E[u_{p,r}^K u_{q,r}^K].
\]
Since $E[u_{i,r}^Q u_{j,r}^Q] = \sigma_Q^2 (x_i^\top x_j)$ and $E[u_{p,r}^K u_{q,r}^K] = \sigma_K^2 (x_p^\top x_q)$:
\begin{equation}
E[s_{ip}s_{jq}] = \sigma_Q^2 \sigma_K^2 (x_i^\top x_j)(x_p^\top x_q). \label{eq:s_ip_s_jq}
\end{equation}

\subsection{Computation of $E[(s_{ip}-\bar{s}_i)(s_{jq}-\bar{s}_j)]$}
\[
s_{ip}-\bar{s}_i = s_{ip} - \frac{1}{T} \sum_{t=1}^T s_{it}, \quad
s_{jq}-\bar{s}_j = s_{jq} - \frac{1}{T} \sum_{u=1}^T s_{ju}.
\]
Multiply and take expectation:
\[
E[(s_{ip}-\bar{s}_i)(s_{jq}-\bar{s}_j)] = \sigma_Q^2 \sigma_K^2 (x_i^\top x_j) \big[(x_p^\top x_q) - \frac{1}{T} \sum_t x_t^\top x_q - \frac{1}{T} \sum_u x_p^\top x_u + \frac{1}{T^2} \sum_{t,u} x_t^\top x_u \big].
\]
Recognize as centered inner product with mean vector $\mu = \frac{1}{T} \sum_t x_t$:
\begin{equation}
E[(s_{ip}-\bar{s}_i)(s_{jq}-\bar{s}_j)] = \sigma_Q^2 \sigma_K^2 (x_i^\top x_j)((x_p - \mu)^\top (x_q - \mu)). \label{eq:centered_term}
\end{equation}

\subsection{Combine All Terms: $C_0$ and $C_1$}
From \eqref{eq:K_RAP_E} and \eqref{eq:a_ip_a_jq}:
\[
E[K_{ij}] = d_k \sigma_V^2 \sum_{p,q} \left[ \frac{1}{T^2} + \frac{1}{T^2} E[(s_{ip}-\bar{s}_i)(s_{jq}-\bar{s}_j)] \right] (x_p^\top x_q).
\]
\textbf{Constant term:}
\begin{equation}
C_0 = \frac{d_k \sigma_V^2}{T^2} \sum_{p,q} x_p^\top x_q. \label{eq:C0}
\end{equation}
\textbf{Linear kernel term:} Using \eqref{eq:centered_term}:
\begin{equation}
C_1 = \frac{d_k \sigma_V^2 \sigma_Q^2 \sigma_K^2}{T^2} \sum_{p,q} ((x_p-\mu)^\top (x_q-\mu)) (x_p^\top x_q). \label{eq:C1}
\end{equation}

\subsection{Final RAPK Expectation}
\begin{equation}
E[K_\text{RAP}]_{ij} \approx C_0 + C_1 (x_i^\top x_j). \label{eq:RAPK_final}
\end{equation}
Matrix form:
\begin{equation}
E[K_\text{RAP}] \approx C_0 \mathbf{11}^\top + C_1 XX^\top.
\end{equation}
If the input is centered ($\mu=0$), then $S_2 = \sum_{p,q} (x_p^\top x_q)^2 = \|XX^\top\|_F^2$ and
\[
C_1 = \frac{d_k \sigma_V^2 \sigma_Q^2 \sigma_K^2}{T^2} \|XX^\top\|_F^2.
\]

\subsection{Interpretation}
The kernel consists of a dominant constant baseline ($C_0$) and a \textbf{linear similarity term} ($C_1$). The latter depends linearly on $x_i^\top x_j$, demonstrating that the expected RAPK structure effectively acts as a linear kernel shifted by a global averaging offset in the high-dimensional limit.

\section{Additional Metrics and Implementation Details} \label{app:metrics}

\subsection{Weighted Transition Entropy (WTE)}
Algorithm~\ref{alg:wte} details the computation of weighted transition entropy, which quantifies the unpredictability of transitions in a sleep stage sequence.

\begin{algorithm}[!htb]
\caption{Computation of Weighted Transition Entropy (WTE)}
\label{alg:wte}
\begin{algorithmic}[1]
\Require Sequence of predicted sleep stages $\mathbf{s} = [s_1, \dots, s_T]$, Number of classes $C$
\Ensure Weighted transition entropy $H_\text{weighted}$

\State Initialize transition count matrix $T \in \mathbb{Z}^{C \times C}$ as zeros
\For{$t = 1$ \textbf{to} $T-1$}
    \State $T[s_t, s_{t+1}] \gets T[s_t, s_{t+1}] + 1$
\EndFor

\State Compute row sums $R_c = \sum_{j=1}^{C} T[c,j]$ for each class $c$
\State Initialize transition probability matrix $P \in \mathbb{R}^{C \times C}$ as zeros
\For{$c = 1$ \textbf{to} $C$}
    \If{$R_c > 0$}
        \State $P[c,:] \gets T[c,:] / R_c$
    \EndIf
\EndFor

\State Compute per-class entropy: $H_c = - \sum_{j=1}^{C} P[c,j] \log P[c,j]$ for all $c$ with $R_c > 0$
\State Compute weighted transition entropy:
\[
H_\text{weighted} = \sum_{c=1}^{C} \frac{R_c}{\sum_{c'} R_{c'}} \cdot H_c
\]
\State \Return $H_\text{weighted}$
\end{algorithmic}
\end{algorithm}

\subsection{Local Smoothness Influence Index (LSII)}
Algorithm~\ref{alg:lsii} details the computation of the Local Smoothness Influence Index, which measures the degree to which sequence-level corrections are supported by local temporal consensus.

\begin{algorithm}[!htb]
\caption{Computation of Local Smoothness Influence Index (LSII)}
\label{alg:lsii}
\begin{algorithmic}[1]
\Require Non-sequential predictions $\mathbf{y}^{\text{none}}$, Corrected predictions $\mathbf{y}^{\text{corr}}$, Ground-truth labels $\mathbf{y}^{\text{true}}$, Window size $W$
\Ensure Mean LSII score

\State $T \gets |\mathbf{y}^{\text{none}}|$
\State $\mathcal{I} \gets \{ i \mid y^{\text{none}}_i \neq y^{\text{corr}}_i \}$
\If{$|\mathcal{I}| = 0$}
    \State \Return \textbf{None}
\EndIf

\State Partition the sequence into non-overlapping windows $\{[t, t+W)\}$
\State Initialize empty list $\mathcal{L}$

\ForAll{$i \in \mathcal{I}$}
    \State Identify window $[s,e)$ containing index $i$
    \State $y \gets y^{\text{corr}}_i$
    \State $\mathcal{W} \gets \{ y^{\text{corr}}_j \mid j \in [s,e), j \neq i \}$
    \State $\text{LSII}_i \gets \frac{1}{|\mathcal{W}|} \sum\limits_{j \in \mathcal{W}} \mathbb{I}[y^{\text{corr}}_j = y]$
    \State Append $\text{LSII}_i$ to $\mathcal{L}$
\EndFor

\State \Return $\text{mean}(\mathcal{L})$
\end{algorithmic}
\end{algorithm}

\section{Generality of RT: Integration into Existing SOTA Model}

To demonstrate the generality of RT, we integrate it into a recent strong baseline, AnySleep~\cite{grieger2025anysleepchannelagnosticdeeplearning}, under a strictly controlled protocol. All models are trained from scratch using the official implementation, with identical preprocessing, optimization, and training schedules. The only modification is replacing the original sequence aggregation module (mean pooling) with a frozen RT, leaving all other components unchanged. This setup isolates the effect of RT as a plug-in module. As shown in Table~\ref{tab:anysleep}, RT consistently improves performance over the original AnySleep across multiple datasets. Notably, gains are observed on EDF-20, EDF-ST, EDFX, and SHHS1, demonstrating that RT can be seamlessly integrated into an existing SOTA pipeline and further enhance its performance. This improvement is achieved without additional architectural redesign, highlighting RT as a simple yet effective drop-in replacement for sequence aggregation.
These results confirm that RT is not tied to a specific model design, but functions as a general-purpose sequence-level operator that can be directly incorporated into competitive architectures to improve performance, supporting its broad applicability beyond our proposed framework.

\begin{table}[t]
\centering
\scriptsize
\setlength{\tabcolsep}{3.5pt}
\renewcommand{\arraystretch}{0.95}

\caption{
Generality of RT under integration into a strong baseline (AnySleep) (ACC / Weighted F1 \%). 
\textbf{Top:} original AnySleep model trained from scratch under a controlled protocol. 
\textbf{Middle:} integration of a frozen RT with $d_k=128$. 
\textbf{Bottom:} RT with increased projection dimension ($d_k=512$).
}
\label{tab:anysleep}

\begin{tabular}{lcccc}
\toprule
\textbf{Method} & EDF-20 & EDF-ST & EDFX & SHHS1 \\
\multicolumn{1}{c}{} & \multicolumn{1}{c}{\tiny ACC / F1} &
\multicolumn{1}{c}{\tiny ACC / F1} &
\multicolumn{1}{c}{\tiny ACC / F1} &
\multicolumn{1}{c}{\tiny ACC / F1} \\
\midrule

\rowcolor{gray!20}
AnySleep &
74.63$\pm$\scriptsize6.21 / 71.74$\pm$\scriptsize5.75 &
74.82$\pm$\scriptsize1.68 / 71.51$\pm$\scriptsize2.31 &
81.92$\pm$\scriptsize0.89 / 80.71$\pm$\scriptsize1.18 &
86.89$\pm$\scriptsize0.53 / 86.07$\pm$\scriptsize0.93 \\

\rowcolor{blue!15}
RT ($d_k=128$) &
79.13$\pm$\scriptsize3.41 / 78.19$\pm$\scriptsize4.07 &
74.79$\pm$\scriptsize3.36 / 73.37$\pm$\scriptsize3.78 &
81.94$\pm$\scriptsize0.73 / 81.38$\pm$\scriptsize0.61 &
\textbf{87.04$\pm$\scriptsize0.65 / 86.89$\pm$\scriptsize0.67} \\

\rowcolor{blue!25}
RT ($d_k=512$) &
\textbf{81.64$\pm$\scriptsize1.48 / 80.91$\pm$\scriptsize2.13} &
\textbf{77.41$\pm$\scriptsize1.48 / 77.36$\pm$\scriptsize1.38} &
\textbf{82.80$\pm$\scriptsize0.98 / 81.96$\pm$\scriptsize1.14} &
86.80$\pm$\scriptsize0.43 / 86.48$\pm$\scriptsize0.52 \\

\bottomrule
\end{tabular}
\end{table}

\section{Direct Sequence Modeling on Raw Signals}

To comprehensively validate the intrinsic sequence modeling capability of the Random Transformer, we investigated whether its smoothing effect persists even without a dedicated epoch encoder. In this experiment, we applied sequence modeling directly to raw EEG signals. As presented in Table~\ref{tab:raw_signal}, a baseline MLP operating independently on each epoch performs close to chance level ($22.19\%$ ACC), failing to capture any temporal context. Crucially, incorporating a randomly initialized sequence encoder immediately yields a substantial performance leap ($\sim 18\%$ absolute gain in Accuracy), demonstrating that the RAPK mechanism effectively enforces local contextual smoothing even on raw, noisy input features. While adding positional encoding further stabilizes this behavior, the Random Transformer alone recovers significant sequential dependencies. However, the gap between this setup and the two-stage framework (Table~\ref{tab:baseline}) underscores that while random attention provides robust temporal smoothing, a strong epoch-level feature extractor remains essential for high-performance sleep staging.

\begin{table}[ht]
\centering
\caption{Performance of direct sequence modeling on raw EEG signals (Sleep-EDF-20).}
\label{tab:raw_signal}
\begin{tabular}{lcc}
\toprule
Method & ACC (\%) & Weighted F1 (\%) \\
\midrule
\rowcolor{gray!20}
MLP (Supervised) 
& 22.19 $\pm$ 0.35 
& 23.45 $\pm$ 0.34 \\

\rowcolor{blue!15}
\textbf{Random Transformer + MLP (w/o Positional Encoding)} 
& \textbf{39.86 $\pm$ 1.46} 
& \textbf{33.81 $\pm$ 1.11} \\

\rowcolor{blue!15}
\textbf{Random Transformer + MLP (w/ Positional Encoding)} 
& \textbf{40.22 $\pm$ 1.22} 
& \textbf{34.03 $\pm$ 1.07} \\

Train Transformer + MLP 
& 50.97 $\pm$ 1.55 
& 48.04 $\pm$ 1.87 \\
\bottomrule
\end{tabular}
\end{table}

\section{RAPK Approximation Error}

\begin{figure}[t]
\centering
\includegraphics[width=0.7\columnwidth]{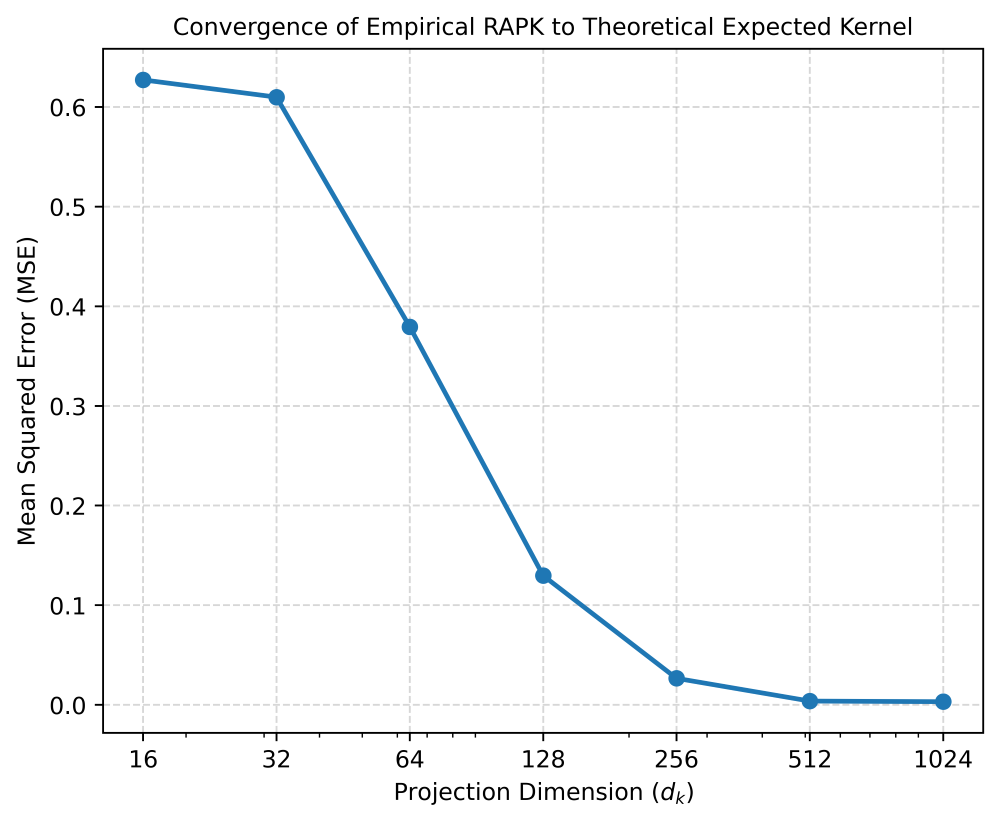}
\caption{
MSE between empirical random Transformer kernels and the closed-form RAPK prediction across varying model widths $d_k$.}
\label{fig:rapk_mse}
\vspace{-5pt}
\end{figure}

\begin{figure}[t]
\centering
\includegraphics[width=\linewidth]{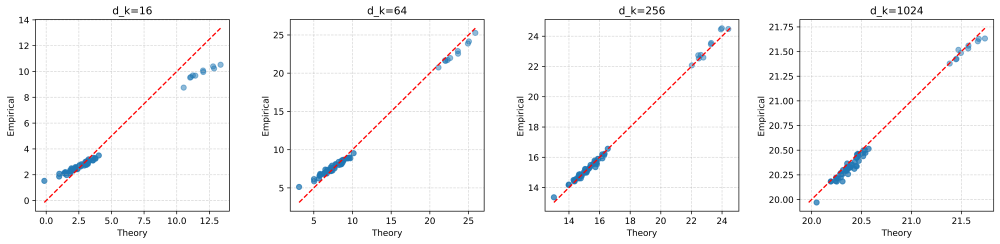}
\caption{
Empirical versus theoretical RAPK kernel values across varying model widths $d_k$.}
\label{fig:rapk_scatter}
\vspace{-5pt}
\end{figure}

\begin{figure}[t]
\centering
\includegraphics[width=\linewidth]{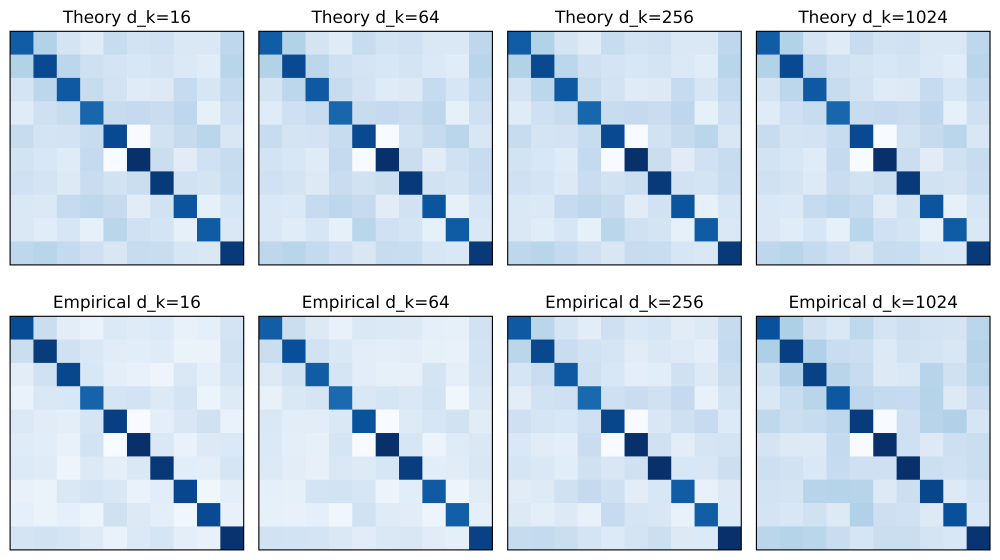}
\caption{
Comparison of empirical random Transformer kernels and the closed-form RAPK prediction.}
\label{fig:rapk_heatmap}
\vspace{-5pt}
\end{figure}

We first quantify the overall approximation gap between the closed-form RAPK and empirical random Transformer (RT) kernels using Monte Carlo simulations (1000 trials) across $d_k \in [16,1024]$. As shown in Fig.~\ref{fig:rapk_mse}, the mean squared error (MSE) decreases monotonically with increasing width and approaches zero at large $d_k$, confirming that RAPK is asymptotically exact. At small dimensions, a finite error remains due to higher-order truncation effects. We then examine the structure of this approximation gap. Despite non-zero MSE at small widths (e.g., $d_k=16$), Fig.~\ref{fig:rapk_scatter} shows that empirical and theoretical kernels maintain a strictly monotonic and highly linear relationship, indicating strong preservation of relative similarity. Furthermore, as shown in Fig.~\ref{fig:rapk_heatmap}, empirical heatmaps closely match theoretical predictions, preserving key structural patterns such as block-diagonal smoothing (state inertia) and clear transition boundaries. These results demonstrate that RAPK is not only asymptotically exact in magnitude but also structurally faithful at practical dimensions, with performance driven primarily by preserved kernel geometry rather than pointwise numerical precision.

\section{Empirical Validation of Attention Logit Concentration}

\begin{figure}[t]
\centering

\begin{subfigure}[b]{0.49\linewidth}
\centering
\includegraphics[width=\textwidth]{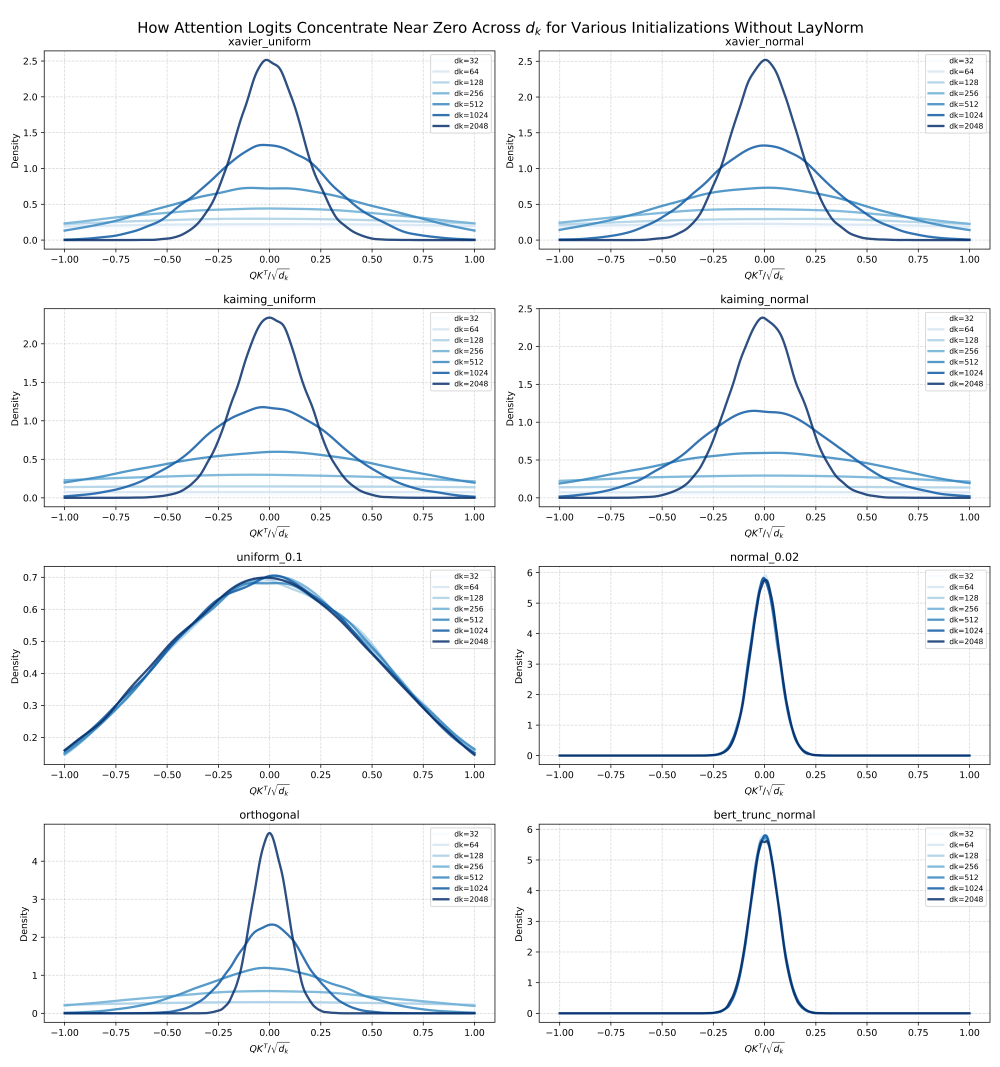}
\caption{Without LayerNorm}
\label{fig:attn_logits_no_ln}
\end{subfigure}
\hfill
\begin{subfigure}[b]{0.49\linewidth}
\centering
\includegraphics[width=\textwidth]{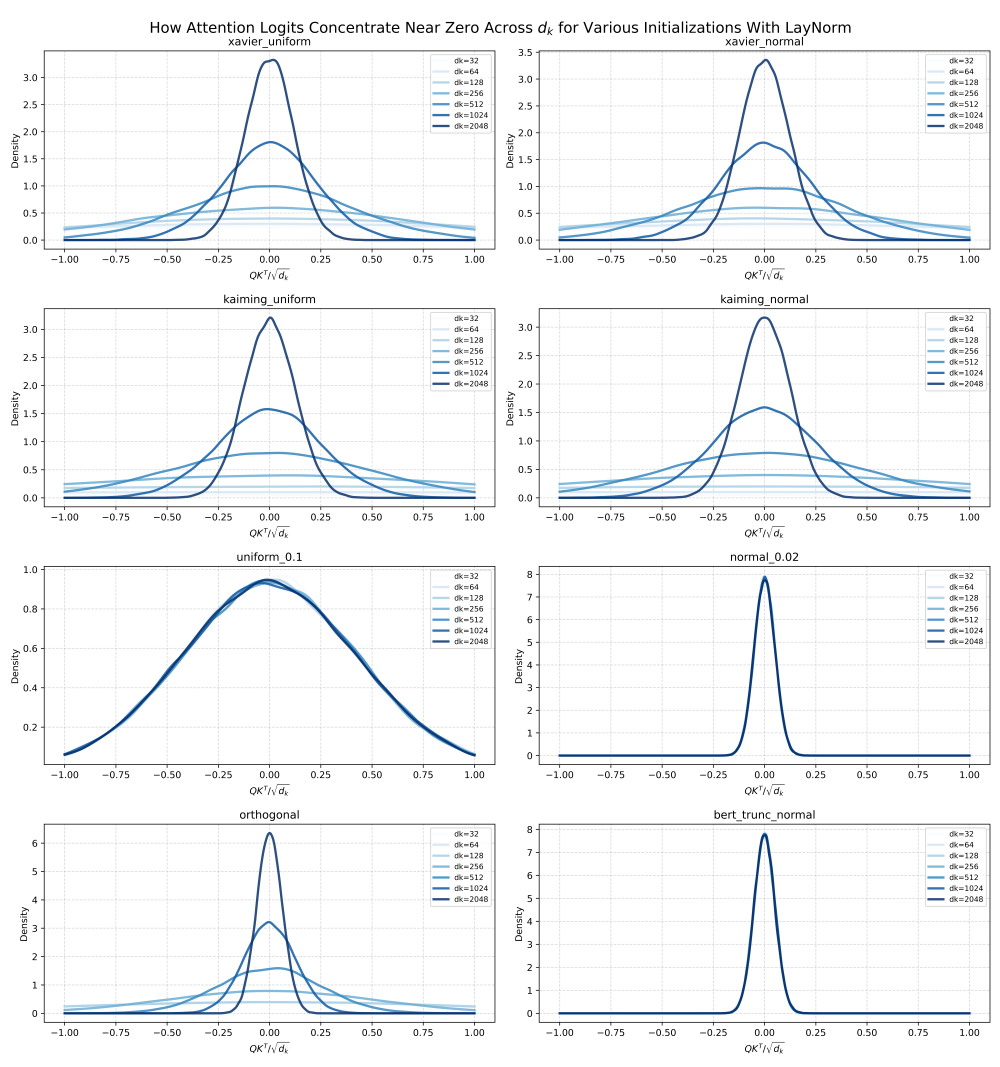}
\caption{With LayerNorm}
\label{fig:attn_logits_ln}
\end{subfigure}

\caption{
Distribution of pre-softmax attention logits with and without Layer Normalization across initialization schemes and varying $d_k$. LayerNorm stabilizes feature statistics and enhances concentration near zero, promoting the near-linear regime of the softmax operation.
}
\label{fig:attn_logits_compare}

\end{figure}

We empirically validate the RAPK assumption that pre-softmax attention logits $QK^\top / \sqrt{d_k}$ concentrate near zero under standard Transformer settings. As shown in Fig.~\ref{fig:attn_logits_no_ln} and Fig.~\ref{fig:attn_logits_ln}, we analyze distributions across 8 initialization schemes and multiple dimensions $d_k \in [32, 2048]$. With Layer Normalization (Fig.~\ref{fig:attn_logits_ln}), logits exhibit significantly sharper concentration around zero due to stabilized feature statistics and bounded input magnitudes. Consequently, most values are confined within a narrow range, placing the softmax in a near-linear regime. Without LayerNorm (Fig.~\ref{fig:attn_logits_no_ln}), distributions are broader, but still show clear concentration-of-measure behavior under standard initializations (e.g., Xavier and Kaiming), where increasing $d_k$ leads to progressive collapse toward zero. Both normalization and initialization jointly drive attention logits into a near-zero regime, confirming that the RAPK assumption holds naturally in practical Transformer implementations.

\section{Correlation Analysis of Sequence Structure Metrics}
\label{app:correlation_metrics}

\begin{figure}[ht]
    \centering
    \includegraphics[width=0.7\columnwidth]{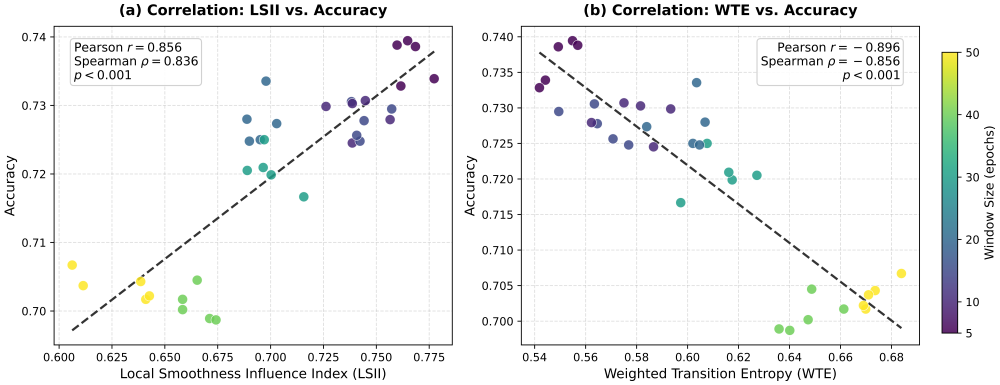}
    
    \caption{Correlation analysis between sequence structure metrics and accuracy across 35 experimental configurations (5 random seeds $\times$ 7 window lengths). The color gradient represents the temporal window size.}
    \vspace{-5pt} 
    \label{fig:metric_correlation}
\end{figure}

To validate LSII and WTE as quantitative proxies for sequence smoothing, we analyzed their correlation with accuracy across 35 experimental configurations on Sleep-EDF-20 (5 seeds $\times$ 7 window lengths $W \in [5, 50]$). Figure~\ref{fig:metric_correlation} reveals strong deterministic links between sequence regularity and model performance.

\textbf{Local Smoothness Influence Index (LSII).} Figure~\ref{fig:metric_correlation}(a) shows a strong positive correlation ($r \approx 0.86, p < 0.001$). Configurations with smaller windows ($W \in [5, 10]$) yield higher smoothness and superior accuracy, confirming that performance gains arise directly from the model's ability to enforce local temporal agreement and correct isolated epoch errors.

\textbf{Weighted Transition Entropy (WTE).} Figure~\ref{fig:metric_correlation}(b) demonstrates a strong negative correlation ($r \approx -0.90, p < 0.001$).Lower entropy consistently aligns with higher accuracy, suggesting that the Random Transformer acts as a temporal stabilizer. By dampening erratic transitions through averaging, it generates sequences that adhere to the intrinsic stability of sleep physiology. These metrics thus quantitatively link the architecture's smoothing inductive bias to classification success.

\section{Component Ablation}
\begin{figure}[htbp]
    \centering
    \includegraphics[width=0.7\columnwidth]{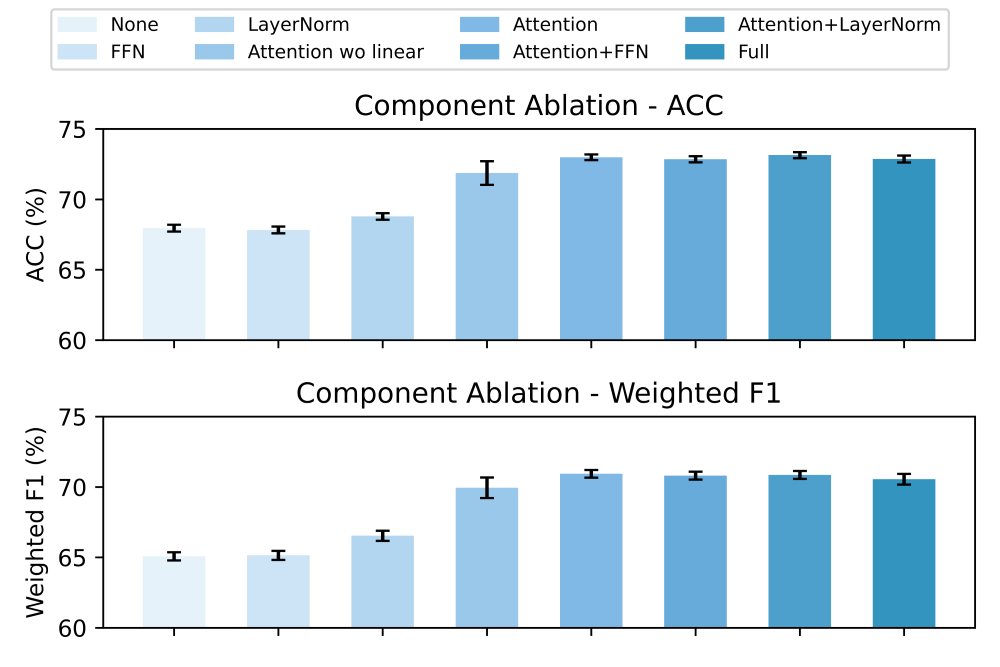}
    \caption{Component ablation on Sleep-EDF-20. Top: ACC (\%); Bottom: Weighted F1 (\%). Colors indicate different module configurations, from left to right: None, FFN, LayerNorm, Attention without linear, Attention, Attention+FFN, Attention+LayerNorm, Full Transformer Encoder.}
    \label{fig:component_ablation}
\end{figure}

We systematically dissected the core modules of the Transformer encoder to identify which components drive the performance gains attributed to the random structural prior. As shown in Figure~\ref{fig:component_ablation}, retaining only the attention mechanism (Only Attention) already yields substantial improvement, demonstrating that the core inductive bias originates from self-attention. From the Random Attention Prior Kernel (RAPK) perspective, random self-attention induces a structured kernel that performs implicit local smoothing, suppressing fluctuations and enhancing sequence consistency. Adding random feedforward network (FFN) and LayerNorm further improves performance by stabilizing feature scales and ensuring the smoothing effect is effective. In contrast, the feedforward network (FFN) contributes minimally under random initialization and can even slightly reduce sequence consistency. Overall, self-attention drives the sequence modeling capability, while random feedforward network (FFN) and LayerNorm provide structural support to form robust smoothing representations without training.

\section{Parameter Sensitivity}
We analyzed the impact of Transformer depth, number of attention heads, initialization schemes, and temporal window lengths to characterize the sensitivity of Random Transformers to architectural complexity and hyperparameter choices.
\paragraph{Layers and Attention Heads}
Our results indicate that a single-layer encoder with eight attention heads achieves the best performance. Increasing the number of heads enhances the model's ability to compute similarity across diverse subspaces, enabling richer representations of multi-scale and complementary sequence dependencies. Under random initialization, this multi-head structure functions as an ensemble of heterogeneous random attention kernels, improving stability and robustness. In contrast, adding additional encoder layers provides no benefit. Each extra layer introduces new random projections and nonlinear transformations that gradually diffuse informative sequence structures and amplify stochastic variations, thereby distorting the underlying sequential geometry. Consequently, deeper random Transformers may weaken the intrinsic sequence smoothness naturally induced by random attention, explaining why increasing \emph{breadth} (heads) is more effective than increasing \emph{depth} (layers) in this setting.

\begin{table}[ht] 
\centering
\caption{Sensitivity of ACC and F1 (\%) to Transformer layers and attention heads (Random mode).}
\label{tab:layer_head}
\begin{tabular}{c|ccc|ccc}
\toprule
\multirow{2}{*}{Layers} & \multicolumn{3}{c|}{ACC} & \multicolumn{3}{c}{Weighted F1} \\
 & 1 head & 4 heads & 8 heads & 1 head & 4 heads & 8 heads \\
\midrule
1 & 73.11 & 73.47 & 73.61 & 71.28 & 71.32 & 71.32 \\
2 & 73.27 & 73.25 & 73.04 & 71.03 & 71.14 & 70.75 \\
3 & 72.98 & 73.36 & 72.95 & 70.76 & 71.40 & 70.79 \\
4 & 73.34 & 73.42 & 72.87 & 71.18 & 71.31 & 70.55 \\
\bottomrule
\end{tabular}%
\end{table}

\paragraph{Initialization Methods} 
We observe that Xavier and Kaiming uniform initializations significantly outperform Gaussian baselines. This divergence is explained by our RAPK framework: Gaussian initialization induces severe feature compression, collapsing the activation variance required for the kernel's similarity term to function ($C_1 \to 0$). This degenerates the model into a blind low-pass filter dominated by the constant term $C_0$. In contrast, uniform initialization preserves feature scale, ensuring attention logits remain in an optimal regime that balances global smoothing with content-adaptive structural retention. Thus, successful random attention requires initialization that maintains the intrinsic geometry of pretrained representations.

\begin{table}[ht]
\centering
\caption{Sensitivity to initialization methods (Random mode).}
\begin{tabular}{lcc}
\toprule
Method & ACC (\%) & Weighted F1 (\%) \\
\midrule
xavier\_uniform & 73.61 $\pm$ 0.14 & 71.32 $\pm$ 0.18 \\
kaiming\_uniform\_relu & 73.54 $\pm$ 0.12 & 71.66 $\pm$ 0.13 \\
uniform\_0.1 & 72.66 $\pm$ 0.32 & 70.24 $\pm$ 0.49 \\
orthogonal & 72.58 $\pm$ 0.06 & 69.96 $\pm$ 0.17 \\
xavier\_normal & 72.23 $\pm$ 0.12 & 69.60 $\pm$ 0.17 \\
kaiming\_normal\_relu & 71.36 $\pm$ 0.26 & 68.82 $\pm$ 0.31 \\
normal\_0.02 & 68.19 $\pm$ 0.27 & 65.22 $\pm$ 0.34 \\
trunc\_normal\_0.02 & 68.05 $\pm$ 0.21 & 65.19 $\pm$ 0.30 \\
\bottomrule
\end{tabular}
\end{table}

\paragraph{Effect of Window Length}
As shown in Figure~\ref{fig:temporal_window_sensitivity}, the performance of the Random Transformer is sensitive to the sequence window size, achieving peak accuracy within a narrow range of 5 to 10 steps. Beyond this range, performance declines steadily. This observation suggests that the Random Transformer primarily functions as a short-range context smoother: within small windows, sleep stages exhibit strong local continuity that random attention can leverage to stabilize predictions. As the window expands, local continuity weakens and more cross-stage transitions occur, making it challenging for random attention to differentiate noise from true transitions, thereby diminishing its smoothing effect and reducing accuracy. This trend is consistent with prior sleep staging studies\cite{9697331,10210638,deng2024lpsgm}.

\section{Visualization Results} \label{app:visualization}

\subsection{Feature Representation Analysis} \label{app:feature_vis}

\begin{figure}[htbp]
    \centering
    \begin{subfigure}[b]{0.24\textwidth}
        \centering
        \includegraphics[width=\textwidth]{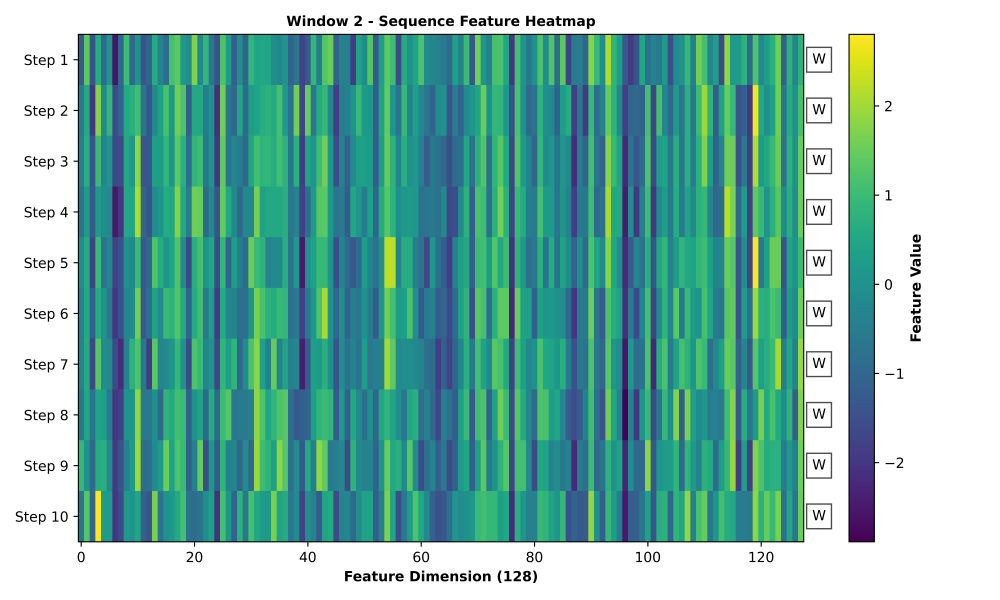}
        \caption{Window 2: Before}
        \label{fig:hm_s1_none}
    \end{subfigure}
    \hfill
    \begin{subfigure}[b]{0.24\textwidth}
        \centering
        \includegraphics[width=\textwidth]{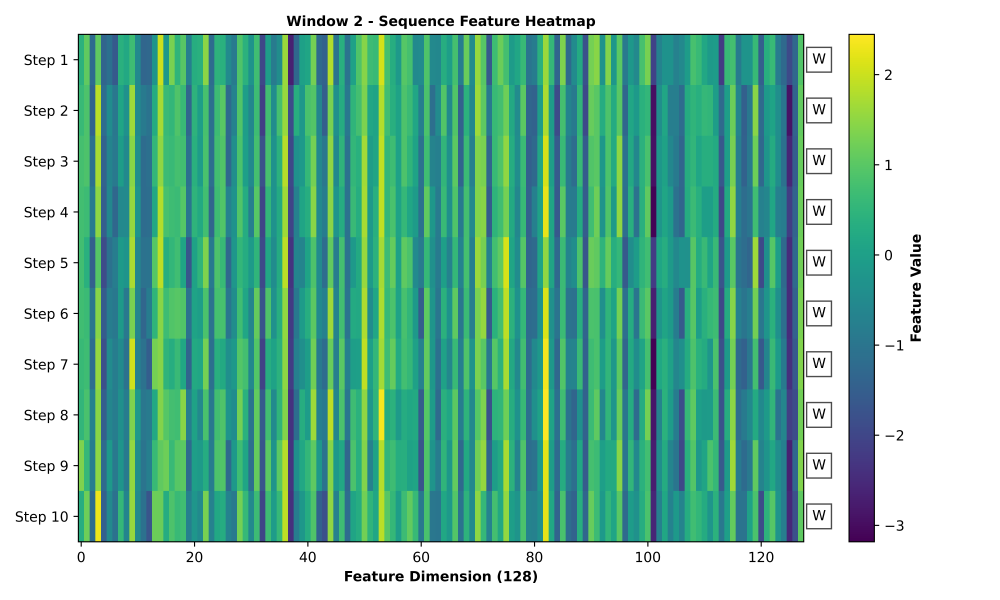}
        \caption{Window 2: After}
        \label{fig:hm_s1_rt}
    \end{subfigure}
    \hfill
    \begin{subfigure}[b]{0.24\textwidth}
        \centering
        \includegraphics[width=\textwidth]{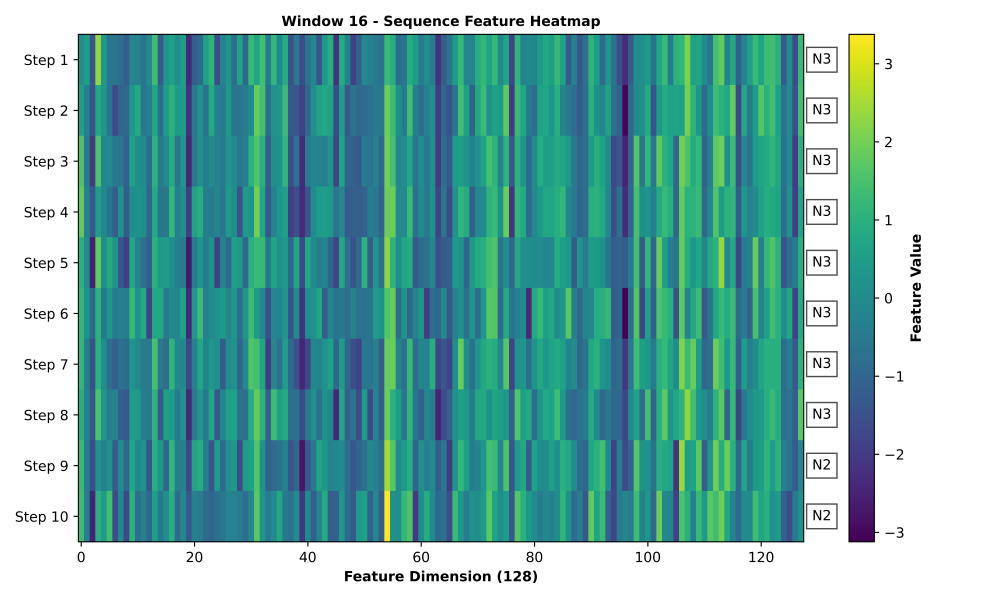}
        \caption{Window 16: Before}
        \label{fig:hm_s15_none}
    \end{subfigure}
    \hfill
    \begin{subfigure}[b]{0.24\textwidth}
        \centering
        \includegraphics[width=\textwidth]{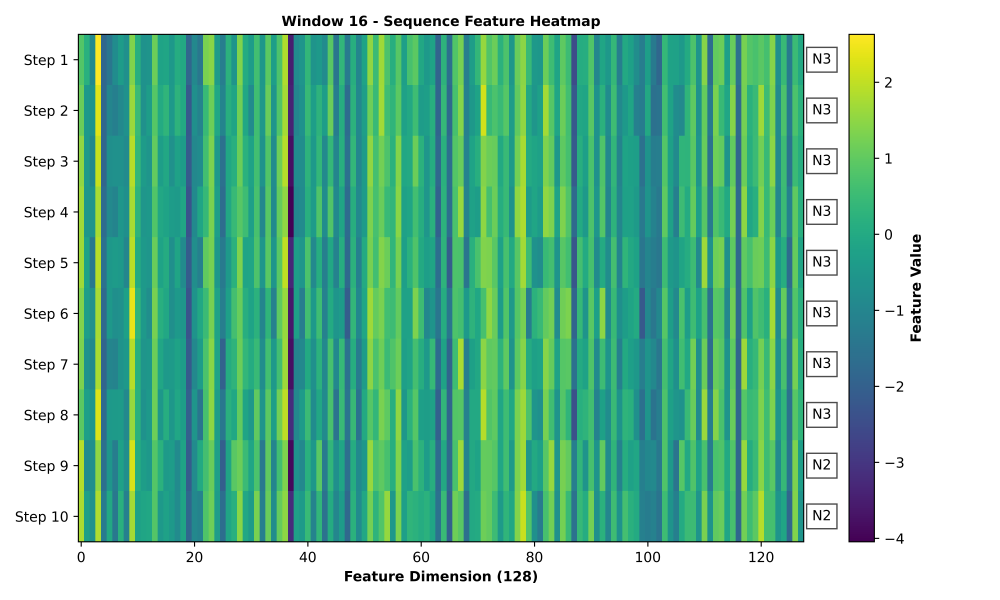}
        \caption{Window 16: After}
        \label{fig:hm_s15_rt}
    \end{subfigure}

    \vspace{0.6em}

    \begin{subfigure}[b]{0.24\textwidth}
        \centering
        \includegraphics[width=\textwidth]{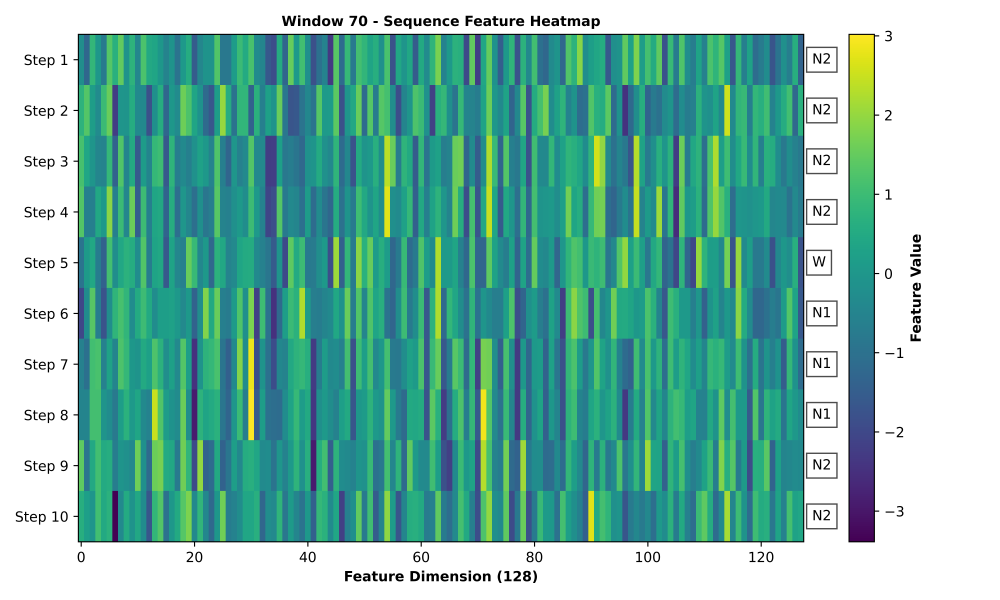}
        \caption{Window 70: Before}
        \label{fig:hm_s69_none}
    \end{subfigure}
    \hfill
    \begin{subfigure}[b]{0.24\textwidth}
        \centering
        \includegraphics[width=\textwidth]{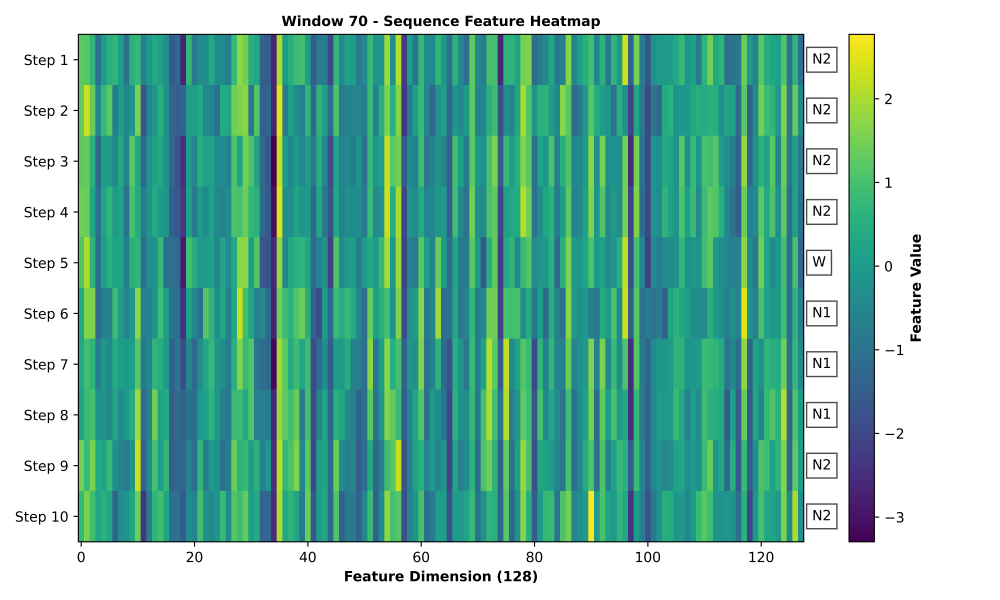}
        \caption{Window 70: After}
        \label{fig:hm_s69_rt}
    \end{subfigure}
    \hfill
    \begin{subfigure}[b]{0.24\textwidth}
        \centering
        \includegraphics[width=\textwidth]{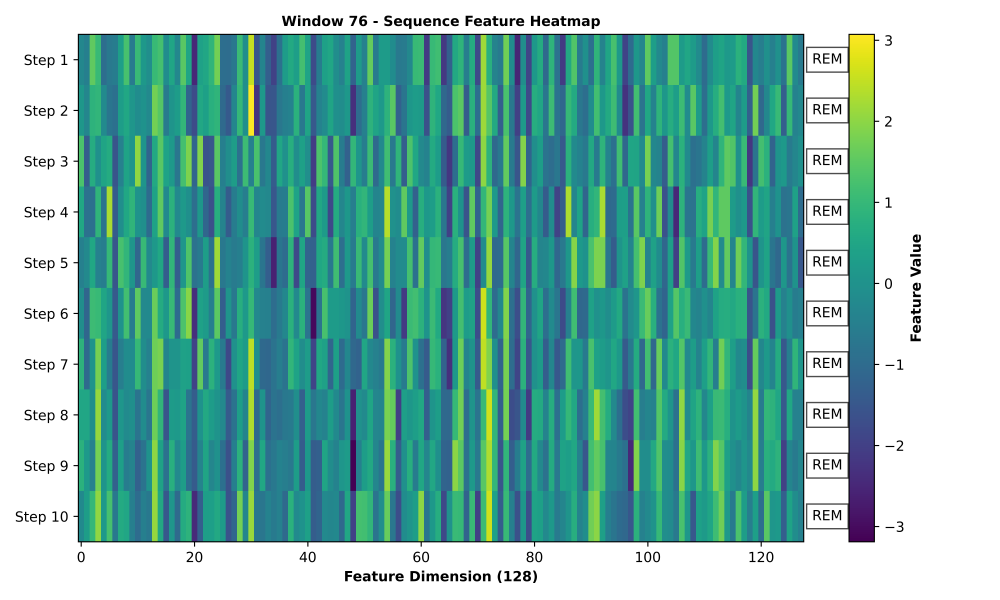}
        \caption{Window 76: Before}
        \label{fig:hm_s75_none}
    \end{subfigure}
    \hfill
    \begin{subfigure}[b]{0.24\textwidth}
        \centering
        \includegraphics[width=\textwidth]{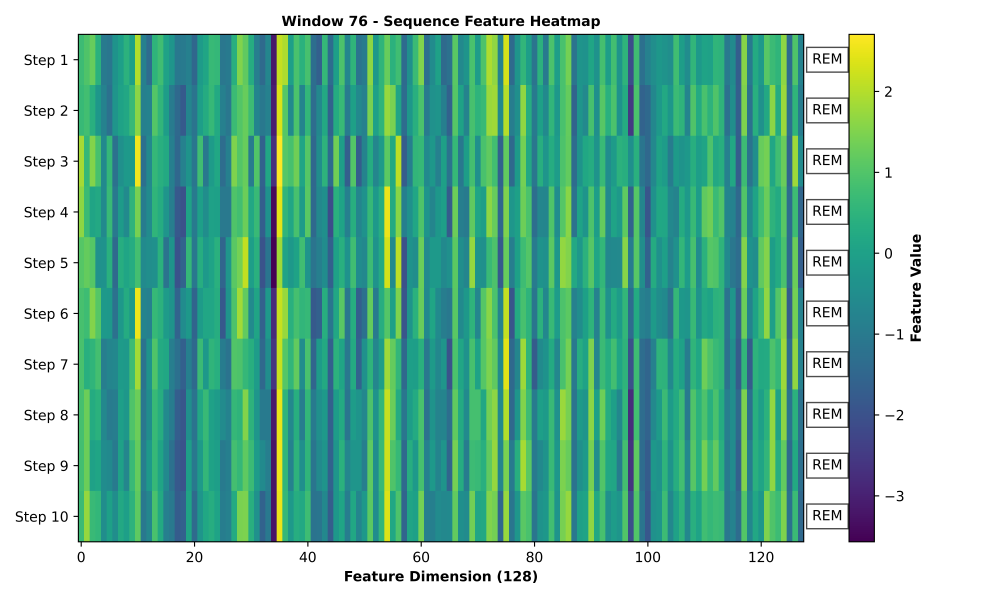}
        \caption{Window 76: After}
        \label{fig:hm_s75_rt}
    \end{subfigure}


    \caption{Visualization of feature heatmaps illustrating the smoothing effect of the Random Transformer.}
    \label{fig:feature_heatmap}
\end{figure}

To investigate the internal mechanism of our proposed method, we visualize the latent feature representations in Figure~\ref{fig:feature_heatmap}. In subject \texttt{SC4181E0} from the Sleep-EDF-20 dataset, we present four representative samples and compare the feature heatmaps produced by the Epoch Encoder with those obtained after processing by the Random Transformer.

As illustrated in the left column (Figs.~\ref{fig:hm_s1_none} and \ref{fig:hm_s15_none} and \ref{fig:hm_s69_none} and \ref{fig:hm_s75_none}), the raw features from the Epoch Encoder exhibit significant high-frequency variance and sequence discontinuities across adjacent time steps, reflecting the noise inherent in independent epoch-level encoding.In stark contrast, after passing through the Random Transformer (right column, Figs.~\ref{fig:hm_s1_rt} and \ref{fig:hm_s15_rt} and \ref{fig:hm_s69_rt} and \ref{fig:hm_s75_rt}), the feature distributions become notably smoother and more homogenized along the temporal axis. This visualization provides direct empirical verification of our RAPK derivation (Eq.~\ref{eq:RAPK_final}). Specifically:
\begin{itemize}
    \item The observed homogenization corresponds to the dominance of the global averaging term ($C_0 \mathbf{1}\mathbf{1}^\top$), which acts as a strong low-pass filter to suppress local fluctuations and noise.
    \item Crucially, the preservation of stage-specific patterns (i.e., the distinct bands remain visible despite smoothing) validates the presence of the linear similarity term ($C_1 XX^\top$). This confirms that the smoothing is content-adaptive rather than blind averaging, allowing the model to enhance sequence coherence while retaining the discriminative structure necessary for classification.
\end{itemize}

\subsection{Visualization of Attention Kernels} 
\label{sec:attn_vis}

To \textbf{demystify} the mechanism of Transformer-based sequence modeling in sleep staging, we visualized the average attention weights within a local window ($W=10$) for both Random and Trained Transformers. This comparison reveals a striking convergence in their functional behavior, suggesting that the "learned" dependencies are functionally similar to the intrinsic bias of the random architecture.

\begin{figure}[ht]
    \centering
    \begin{subfigure}[b]{0.24\textwidth}
        \centering
        \includegraphics[width=\textwidth]{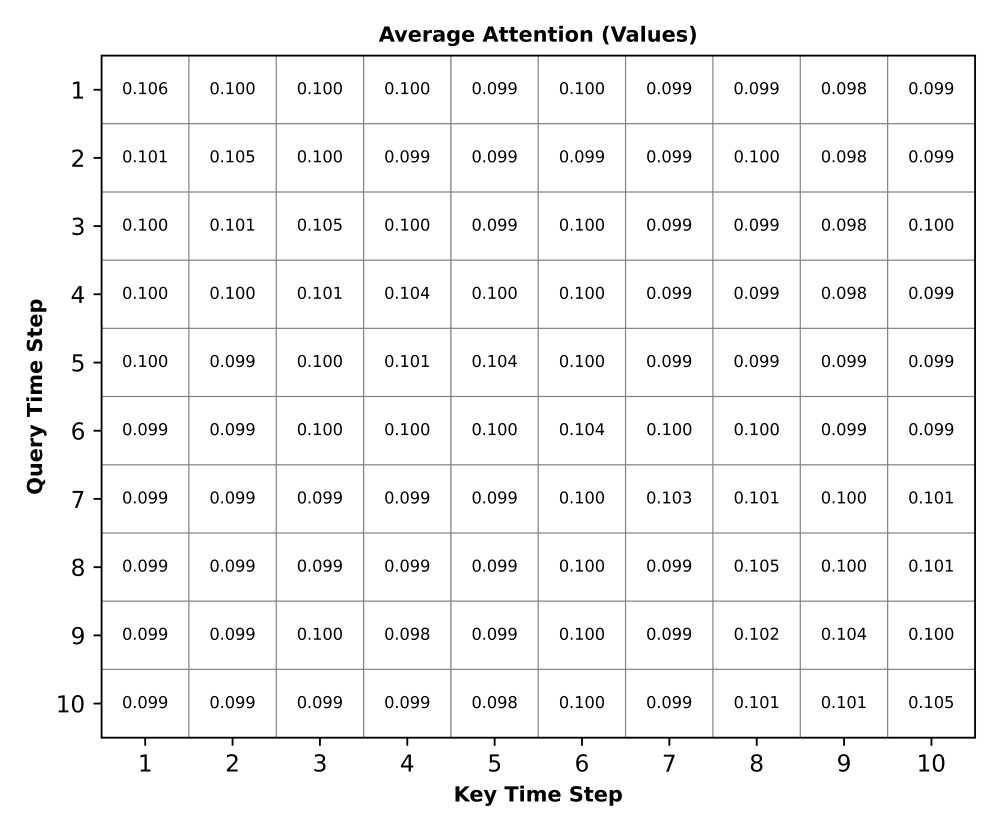}
        \caption{EDF-20}
        \label{fig:attn_edf20_app}
    \end{subfigure}
    \hfill
    \begin{subfigure}[b]{0.24\textwidth}
        \centering
        \includegraphics[width=\textwidth]{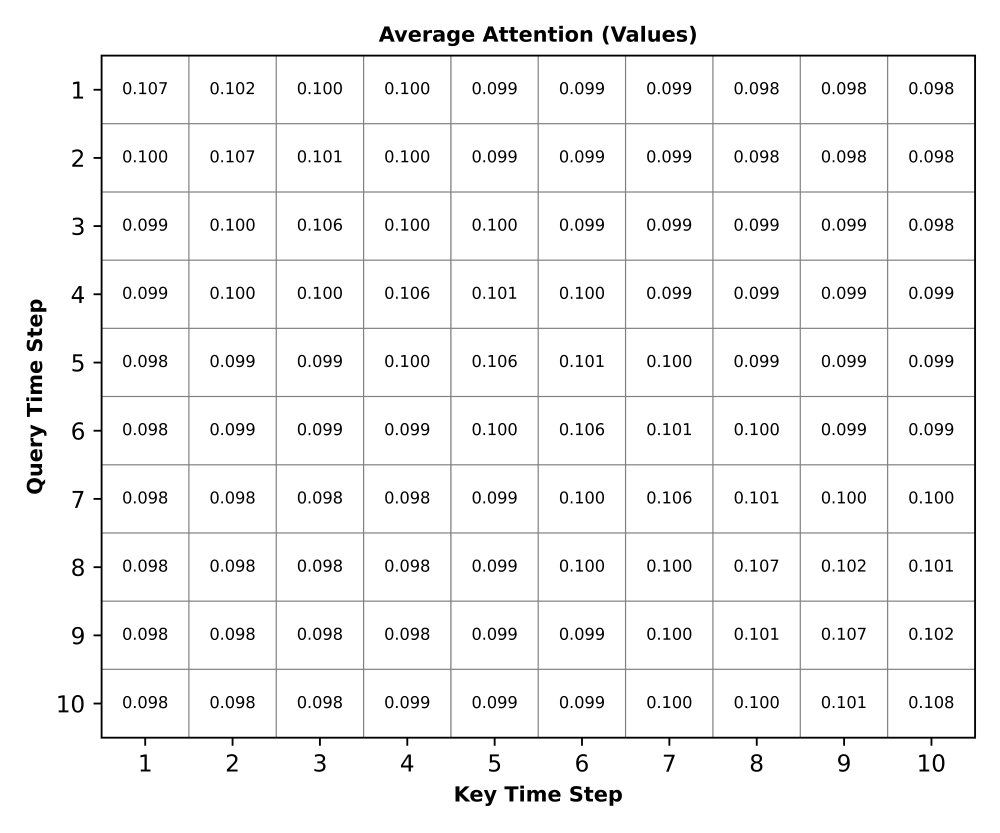}
        \caption{EDFX}
        \label{fig:attn_edfx_app}
    \end{subfigure}
    \hfill
    \begin{subfigure}[b]{0.24\textwidth}
        \centering
        \includegraphics[width=\textwidth]{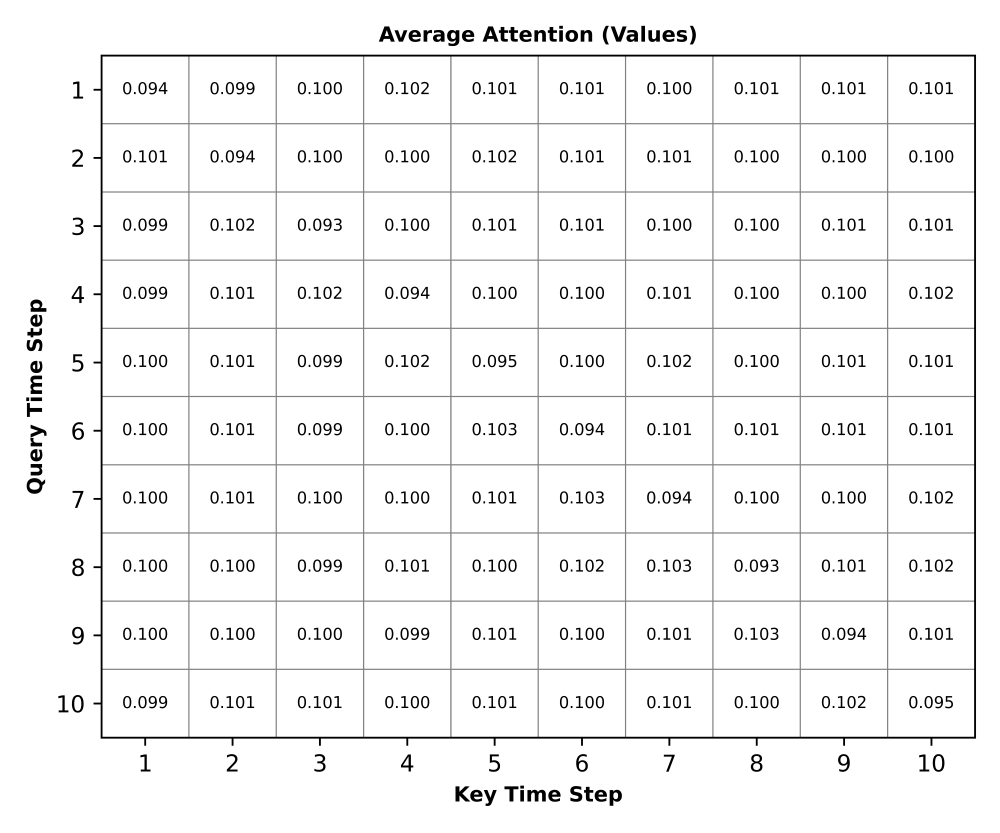}
        \caption{EDF-ST}
        \label{fig:attn_edfst_app}
    \end{subfigure}
    \hfill
    \begin{subfigure}[b]{0.24\textwidth}
        \centering
        \includegraphics[width=\textwidth]{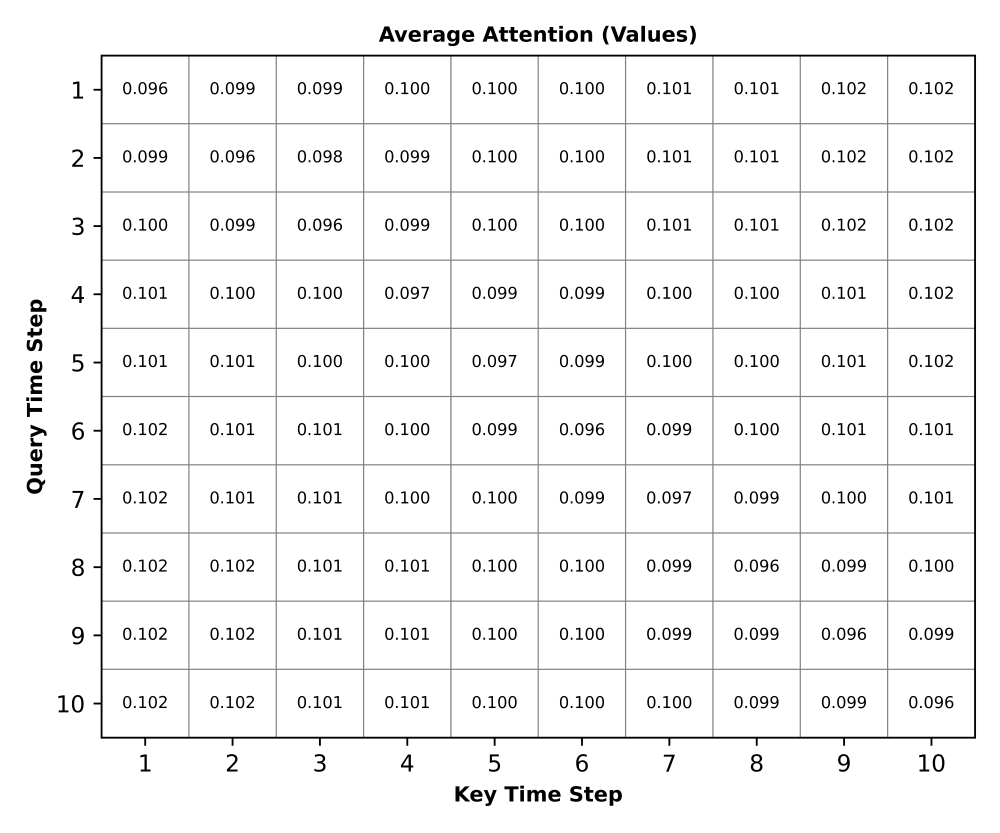}
        \caption{SHHS}
        \label{fig:attn_shhs_app}
    \end{subfigure}

\caption{Average attention weights within a temporal window of size $W=10$ across four datasets using the Random Transformer.}
    \label{fig:attn_weights_random}
\end{figure}

\textbf{Random Transformer} 
As illustrated in Figure~\ref{fig:attn_weights_random}, the weights of the Random Transformer across all datasets converge to approximately $0.1$ (i.e., $1/W$), forming a near-uniform distribution. This empirical observation aligns perfectly with the \textbf{global averaging term ($C_0$)} predicted by our RAPK derivation, where the expected attention score is dominated by the zeroth-order term $E[a_{ip}] \approx 1/T$. This confirms that the Random Transformer inherently functions as a context-smoothing operator, suppressing high-frequency fluctuations by aggregating representations across the temporal window.

\begin{figure}[ht]
    \centering
    \begin{subfigure}[b]{0.24\textwidth}
        \centering
        \includegraphics[width=\textwidth]{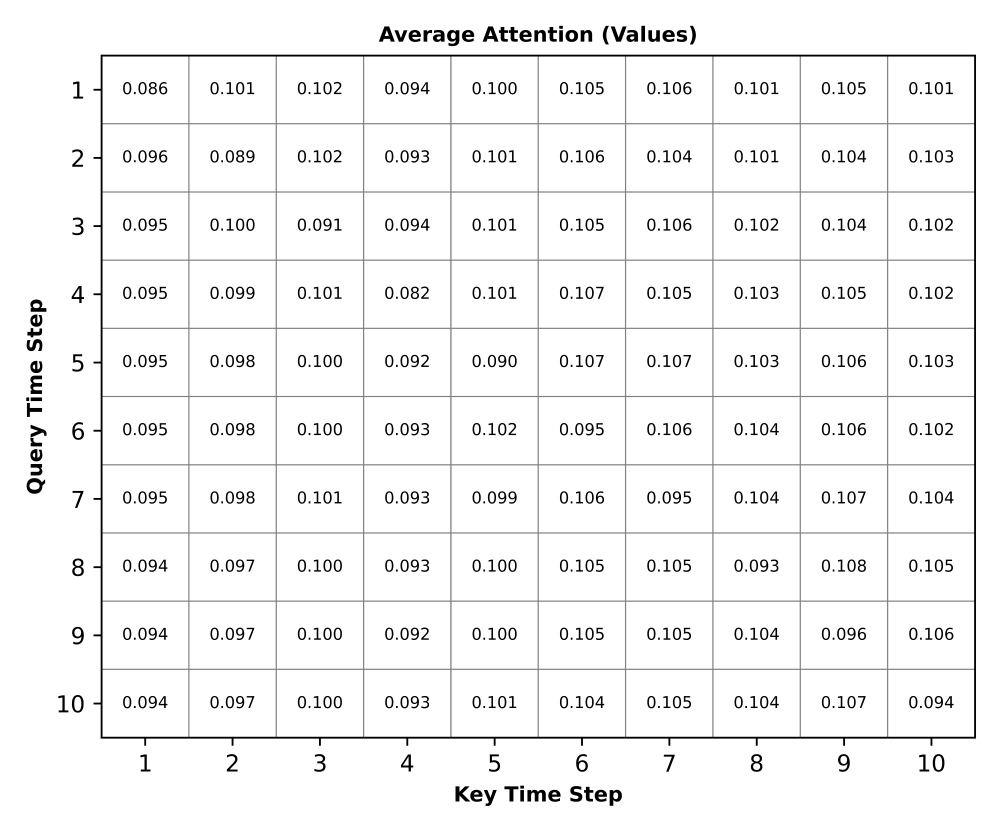}
        \caption{EDF-20}
        \label{fig:attn_edf20_app}
    \end{subfigure}
    \hfill
    \begin{subfigure}[b]{0.24\textwidth}
        \centering
        \includegraphics[width=\textwidth]{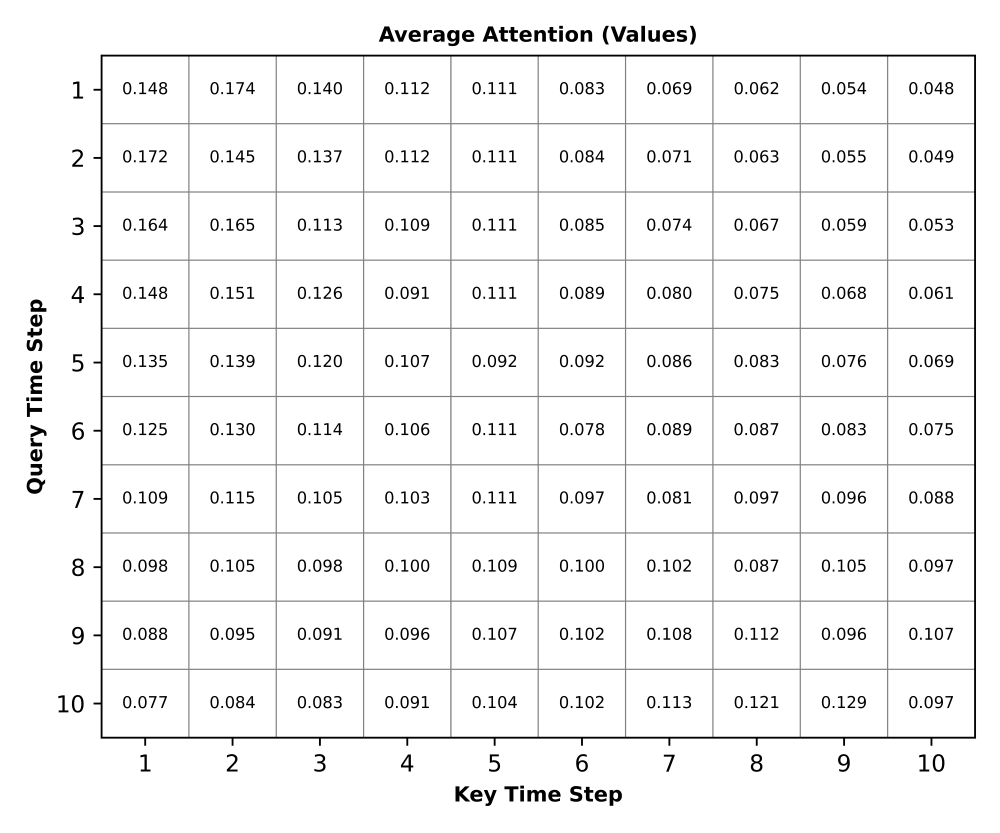}
        \caption{EDFX}
        \label{fig:attn_edfx_app}
    \end{subfigure}
    \hfill
    \begin{subfigure}[b]{0.24\textwidth}
        \centering
        \includegraphics[width=\textwidth]{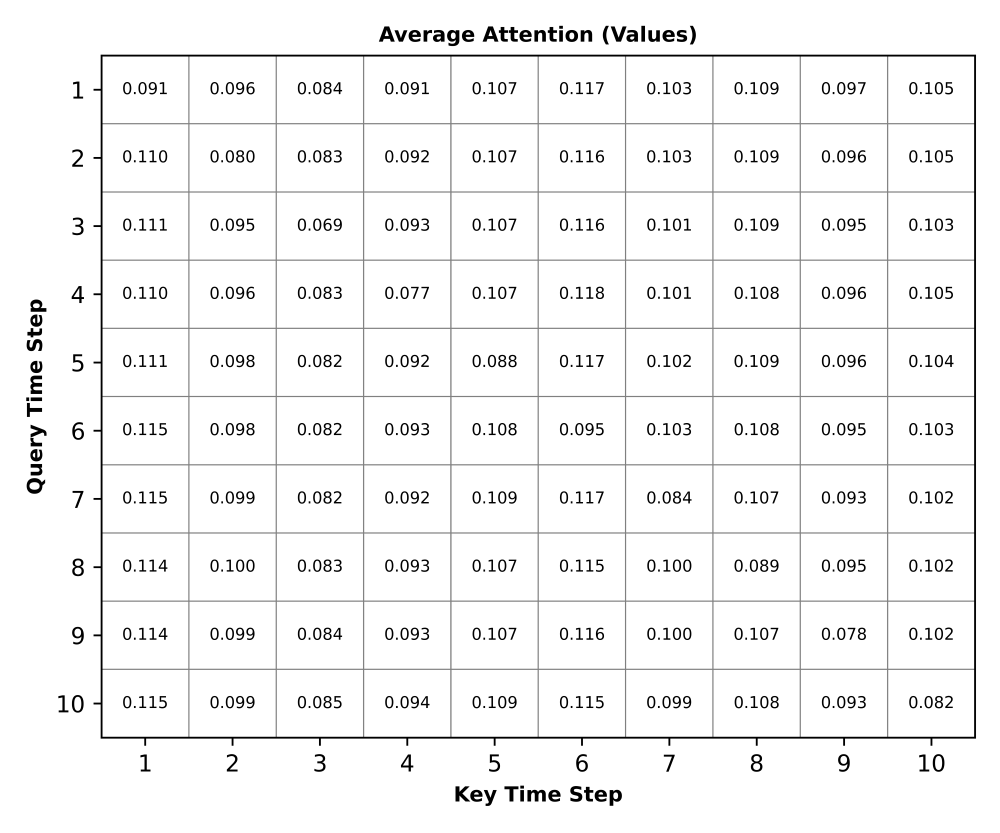}
        \caption{EDF-ST}
        \label{fig:attn_edfst_app}
    \end{subfigure}
    \hfill
    \begin{subfigure}[b]{0.24\textwidth}
        \centering
        \includegraphics[width=\textwidth]{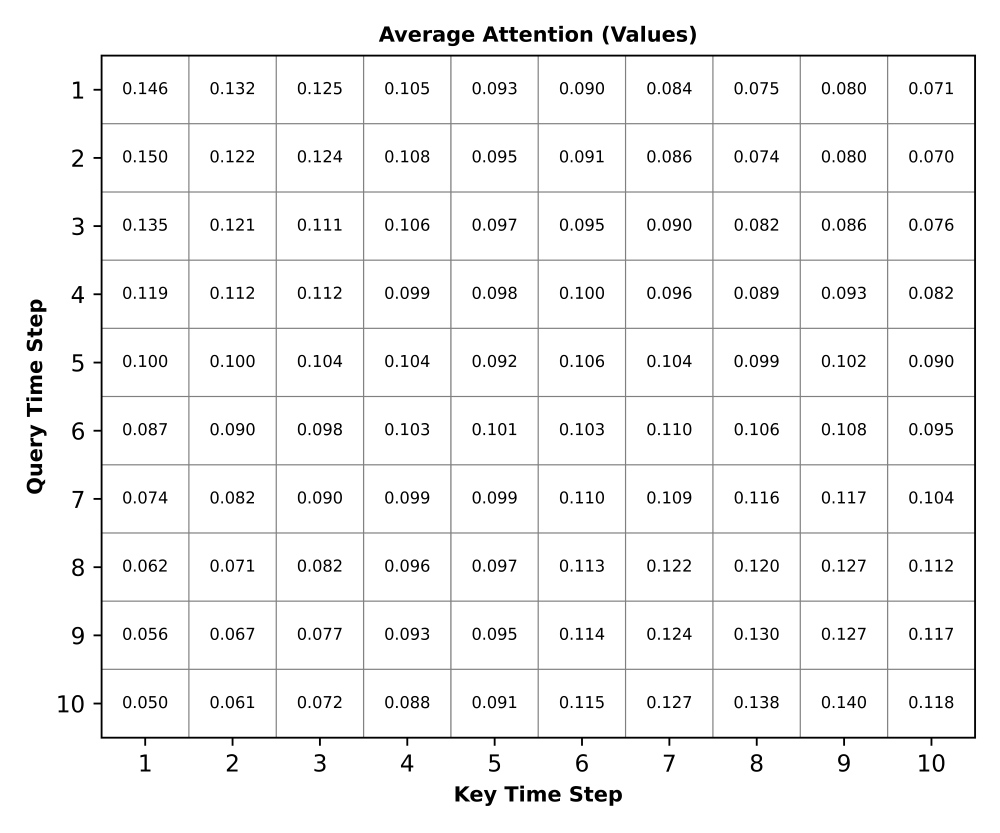}
        \caption{SHHS}
        \label{fig:attn_shhs_app}
    \end{subfigure}

\caption{Average attention weights within a temporal window of size $W=10$ across four datasets using the Train Transformer.}
    \label{fig:attn_weights_trained}
\end{figure}

\textbf{Trained Transformer} 
Does supervised training evolve beyond this simple smoothing to capture complex long-range dependencies? Figure~\ref{fig:attn_weights_trained} visualizes the learned attention kernels of the fully trained models. Surprisingly, we observe no sparse, long-range "skip" connections. Instead, the learned patterns fall into two categories of smoothers:
\begin{itemize}
    \item \textbf{Global Averaging Regime:} In datasets such as Sleep-EDF-20 and Sleep-EDF-ST, the learned weights remain nearly uniform ($\approx 0.1$), functionally identical to the Random Transformer.
    \item \textbf{Local Smoothing Regime:} In SHHS and Sleep-EDFX, the weights exhibit a diagonal-dominant decay, effectively acting as a Gaussian smoothing filter that prioritizes proximal context.
\end{itemize}

These visualizations provide white-box evidence that sleep staging does not require complex dependency modeling. Since the optimization target of a fully trained Transformer effectively collapses into a smoothing operator (either global or local), the Random Transformer's intrinsic averaging bias provides a highly effective "warm start," explaining its ability to rival supervised models without any parameter updates.

\subsection{Feature Activation Density Analysis via RAPK Dynamics} \label{app:feature_actuivation}

\begin{figure}[t]
    \centering
    \includegraphics[width=0.70\columnwidth]{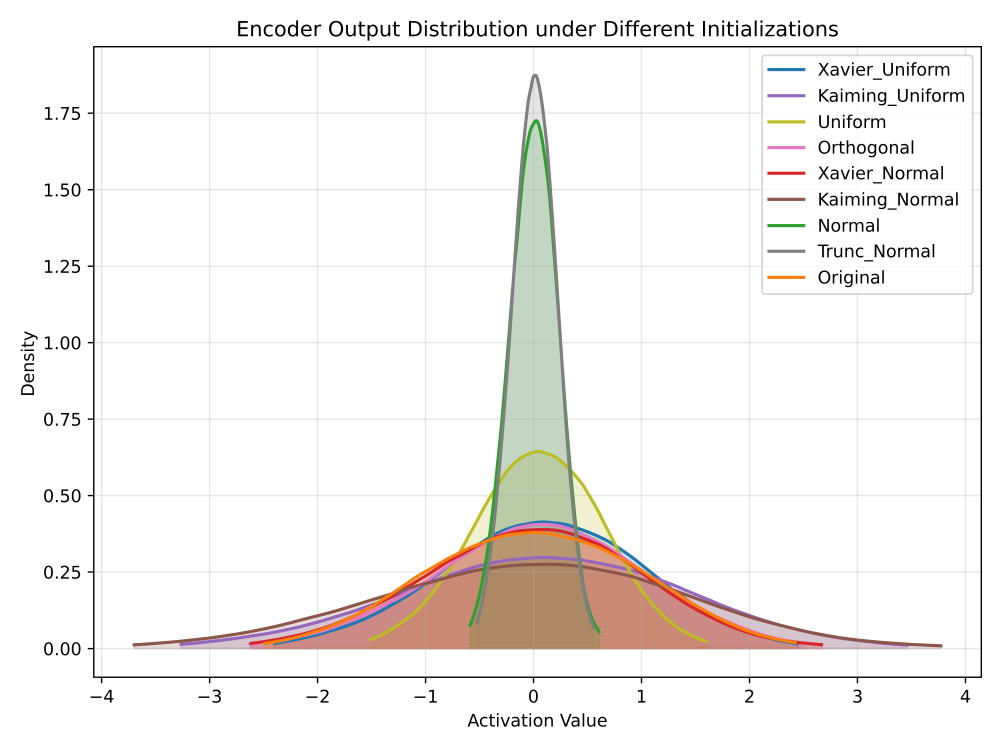}
    \caption{Activation distribution.}
    \label{fig:activation_dist}
\end{figure}

We analyze the impact of initialization schemes on the feature space by synthesizing the empirical evidence from Figure~\ref{fig:activation_dist} with our Random Attention Prior Kernel (RAPK) derivation (Eq.~\ref{eq:RAPK_final}).

\paragraph{The Compressive Effect of Gaussian Initialization.}
As visually corroborated in Figure~\ref{fig:activation_dist}, Standard Normal (Green) and Truncated Normal (Grey) initializations exhibit extremely narrow, high-peaked distributions concentrated around zero. Compared to the Original feature distribution (Orange), these schemes induce severe feature compression.
Theoretically, this collapse in activation variance causes the attention logits $s_{ip}$ to vanish towards zero. In the context of RAPK, this leads to the rapid decay of the linear similarity coefficient ($C_1 \propto \sigma_Q^2 \sigma_K^2$), while the constant term $C_0$ remains dominant. Consequently, the kernel degenerates into a global averaging operator ($E[K_{\text{RAP}}] \approx C_0 \mathbf{1}\mathbf{1}^\top$). This blind over-smoothing erases the distinct geometric structure of the input sequence ($XX^\top$), explaining why Gaussian-initialized models fail to capture meaningful temporal dependencies.

\paragraph{Variance Preservation via Uniform Initialization.}
In stark contrast, the Xavier Uniform (Blue) and Kaiming Uniform (Purple) distributions almost perfectly overlap with the Original distribution (Orange), maintaining the native scale and variance of the input features.
From the RAPK perspective, this variance preservation ensures that the attention scores $s_{ip}$ operate in an optimal linear regime: sufficiently small to satisfy the Taylor expansion approximation, yet significant enough to maintain a non-negligible magnitude for $C_1$. This balance allows the Random Transformer to function as a structure-preserving smoother, where the global term $C_0$ suppresses high-frequency noise while the linear term $C_1$ retains the essential stage-transition boundaries necessary for accurate classification.

\begin{table}[htbp]
\centering
\caption{Detailed information for each of the datasets after preprocessing}
\label{tab:dataset_info}
\begin{tabular}{l c l l l r}
\toprule
\textbf{Datasets} & \textbf{\#Subjects} & \textbf{Channel} & \textbf{Sampling Rate} 
    & \textbf{Stage} & \textbf{Epochs (\%)} \\
\midrule
\multirow{5}{*}{Sleep-EDF-20} & \multirow{5}{*}{20} & \multirow{5}{*}{Fpz-Cz} & \multirow{5}{*}{100 Hz} 
    & W   & 8\,285   (19.6\%)  \\
    & & & & N1  & 2\,804   (6.6\%)   \\
    & & & & N2  & 17\,799  (42.1\%)  \\
    & & & & N3  & 5\,703   (13.5\%)  \\
    & & & & REM & 7\,717   (18.2\%)  \\[3pt]
\multirow{5}{*}{Sleep-EDFX} & \multirow{5}{*}{78} & \multirow{5}{*}{Fpz-Cz} & \multirow{5}{*}{100 Hz} 
    & W   & 65\,951  (33.7\%)  \\
    & & & & N1  & 21\,522  (11.0\%)  \\
    & & & & N2  & 69\,132  (35.4\%)  \\
    & & & & N3  & 13\,039  (6.7\%)   \\
    & & & & REM & 25\,835  (13.2\%)  \\[3pt]
\multirow{5}{*}{Sleep-EDF-ST} & \multirow{5}{*}{22} & \multirow{5}{*}{Fpz-Cz} & \multirow{5}{*}{100 Hz} 
    & W   & 4\,203   (9.9\%)   \\
    & & & & N1  & 3\,653   (8.6\%)   \\
    & & & & N2  & 19\,851  (46.7\%)  \\
    & & & & N3  & 6\,415   (15.1\%)  \\
    & & & & REM & 8\,349   (19.7\%)  \\[3pt]
\multirow{5}{*}{SHHS} & \multirow{5}{*}{329} & \multirow{5}{*}{C4-A1} & \multirow{5}{*}{125 Hz} 
    & W   & 46\,319  (14.3\%)  \\
    & & & & N1  & 10\,304  (3.2\%)   \\
    & & & & N2  & 142\,125 (43.7\%)  \\
    & & & & N3  & 60\,153  (18.5\%)  \\
    & & & & REM & 65\,953  (20.3\%)  \\
\bottomrule
\end{tabular}
\end{table}

\section{Datasets Details}
\label{app:datasets}  
\paragraph{Sleep-EDF:} The Sleep-EDF dataset contains 197 whole-night polysomnography (PSG) recordings, providing EEG (Fpz-Cz, Pz-Cz at 100 Hz), EOG, EMG, etc. Sleep stages are manually annotated by experts according to the R\&K standard (W, S1, S2, S3, S4, REM, MOVEMENT, UNKNOWN)\cite{doi:10.1161/01.CIR.101.23.e215}. Three subsets are used:

\begin{itemize}
    \item \textbf{Sleep-EDF-20:} 20 healthy Caucasian subjects, two nights each; one record missing, total 39 PSG recordings.
    \item \textbf{Sleep-EDFX:} 78 subjects, two nights each, total 153 PSG recordings after excluding missing data.
    \item \textbf{Sleep-EDF-ST:} 22 subjects recorded under temazepam and placebo conditions for two nights each, total 44 PSG recordings.
\end{itemize}

All MOVEMENT and UNKNOWN epochs were excluded, and S3/S4 were merged into N3 according to AASM standards. Only Fpz-Cz EEG channel was used.  

\paragraph{SHHS:} The Sleep Heart Health Study (SHHS) is a multi-center cohort\cite{10.1093/sleep/20.12.1077}. We selected 329 subjects with low apnea-hypopnea index (\text{AHI} \textless 5) and relatively normal sleep patterns~\cite{7446264}. The electroencephalogram (EEG) signal from the C4-A1 channel, originally sampled at 125 Hz, was downsampled to 100 Hz.

\section{Broader impact}
\label{app:Impact Statement}
This work highlights that strong performance in sleep-stage sequence modeling may arise from structural priors rather than extensive training, suggesting opportunities for more efficient and interpretable physiological AI systems. These insights may benefit real-time, resource-constrained, or edge-device applications in sleep monitoring and modulation. At the same time, the findings are context-dependent and should not be overgeneralized to complex clinical decision-making without rigorous validation, as inappropriate deployment could lead to misleading or unsafe outcomes.


\end{document}